\documentclass[10pt,journal,compsoc]{IEEEtran}

\usepackage{times}
\usepackage{epsfig}
\usepackage{graphicx}
\usepackage{amsmath}
\usepackage{amssymb}
\usepackage{times}
\usepackage{helvet}
\usepackage{courier}
\usepackage{graphicx}
\usepackage{booktabs}       
\usepackage{amsfonts}       
\usepackage{amsmath}
\usepackage{amssymb}
\usepackage{url}
\usepackage{multirow}
\usepackage{color}
\definecolor{lightgray}{gray}{0.9}
\definecolor{grey1}{gray}{0.8}
\usepackage[dvipsnames]{xcolor}
\usepackage{color, colortbl}
\usepackage{rotating}
\usepackage{adjustbox}
\begin{document}
%
%
\title{DeepActsNet: Spatial and Motion features from Face, Hands, and Body Combined with Convolutional and Graph Networks for Improved Action Recognition}
\author{Umar Asif,~\IEEEmembership{Senior Member,~IEEE,} Deval Mehta, Stefan Von Cavallar, Jianbin Tang, Stefan Harrer,~\IEEEmembership{Senior Member,~IEEE,}
\thanks{IBM Research, SouthBank, Victoria, Melbourne, Australia, e-mail:umar.asif@gmail.com}}
%
%
%
%
%
%
\maketitle
\begin{abstract}
   Existing action recognition methods mainly focus on joint and bone information in human body skeleton data due to its robustness to complex backgrounds and dynamic characteristics of the environments. In this paper, we combine body skeleton data with spatial and motion features from face and two hands, and present ``Deep Action Stamps (DeepActs)", a novel data representation to encode actions from video sequences. We also present ``DeepActsNet", a deep learning based ensemble model which learns convolutional and structural features from Deep Action Stamps for highly accurate action recognition. Experiments on three challenging action recognition datasets (NTU60, NTU120, and SYSU) show that the proposed model trained using Deep Action Stamps produce considerable improvements in the action recognition accuracy with less computational cost compared to the state-of-the-art methods.
\end{abstract}
%
%
\section{Introduction}
Skeleton-based action recognition has received considerable research focus because of its robustness to dynamic characteristics of real-world environments such as different lighting conditions, variable camera viewpoints, background clutter, and variation in body scales. Existing skeleton-based methods have explored action recognition using both hand-crafted features \cite{fernando2015modeling} as well as deep-learning based features \cite{liu2020disentangling,shi2019skeleton}. In the context of deep learning for action recognition, earlier methods \cite{si2019attention,liu2017skeleton} used Recurrent Neural Networks (RNNs) and Long Short Term Memory Networks (LSTMs) to learn temporal features from the time-series skeleton data. Later, CNN-based methods \cite{feichtenhofer2019slowfast,caetano2019skelemotion} proposed to encode skeleton data into images and employ 2D/3D CNN models for action recognition.  
Recently, graph-based methods such as \cite{shi2019skeleton} and \cite{liu2020disentangling} represented the skeleton data as directed acyclic graphs with joints as vertexes and bones as edges. These graph-based methods learn features based on information in adjacent joints and bones as well as their dependencies and produce state-of-the-art performance on large scale action recognition datasets such as NTU60 \cite{shahroudy2016ntu} and NTU120 \cite{liu2019ntu}.
Most of the existing methods focus on using limited skeleton information based on a pre-defined physical structure of the human body, which may not be optimal for the action recognition task. For instance, the hands may have strong dependencies in recognizing certain action classes such as ``clapping", ``typing" or ``writing". Similarly, facial features can be important for recognizing actions involving facial gestures/deformations such as ``yawning", ``sneeze" or ``wipe face". 
In this paper, we investigate the use of multi-modal information in terms of spatial and motion features extracted from face, body, and the fingers of two hands for action recognition. We also explore ensembles of convolutional and graph networks for learning multi-modal feature representations for action recognition.
In summary, the main contributions of this paper are follows:
\begin{enumerate}
	\item We present ``Deep Action Stamps (DeepActs)", a novel data representation which encodes actions in terms of spatial and motion information extracted from face, hands, and body. To the best of our knowledge, this is the first work which models the spatial and temporal dependencies between facial joints, hand joints, and body joints for action recognition.
	\item We present DeepActsNet, an ensemble of Enhanced Convolutional Graph Networks (ECGN) that learn convolutional and structural features from different feature channels of Deep Actions Stamps. We also develop a lightweight strong baseline, which is morepowerful than the previous methods in terms of recognition accuracy and computational efficiency.
	\item We present ablation studies in terms of the benefits of combining spatial and motion information from face, hands, and body, and the significance of ensembling convolutional and structural features for improving accuracy of challenging action classes. Experiments on three public datasets show that our contributions consistently exceed the state-of-the-art performance on all datasets with considerable margins.
\end{enumerate}
\section{Related Work}
Earlier approaches to action recognition focused on using hand-crafted features such as pairwise position of joints \cite{wang2013learning}, spatial orientation of pairwise joints \cite{jin2012essential}, and statistics-based features \cite{hussein2013human}. However, these methods ignored the semantic connectivity of the human body. 
Later, methods such as \cite{du2015hierarchical,liu2017skeleton} and \cite{zhu2016co,si2019attention} focused on using RNNs and LSTMs for learning spatio-temporal features for action recognition, respectively. Other methods such as \cite{zhang2019view} and \cite{li2018co} presented a view adaptive model and a hierarchical CNN model to learn spatial and temporal features for action recognition, respectively.
Another stream of work encodes skeleton joints information into 2D images and then feed the images into popular CNN models such as ResNet to learn features for action recognition. For instance, the methods of \cite{li2019learning,liu2017enhanced} proposed shape-based visual representations of body skeleton data. The method of \cite{caetano2019skelemotion} proposed a visual representation based on the magnitude and orientation values of skeleton joints. The methods of \cite{feichtenhofer2016convolutional,simonyan2014two} used two-stream ConvNets to fuse skeleton and optical flow information. The method of \cite{li2019temporal} proposed Temporal Bilinear Networks to learn temporal dependencies between joints for action recognition. The work of \cite{jiang2019stm} combined motion modeling into spatio-temporal feature learning. 
Other works in this category utilize 3D CNNs to learn spatio-temporal features. For instance, the work of \cite{xie2017aggregated} used a 3D version of ResNeXt. The Slow-Fast network of \cite{feichtenhofer2019slowfast} used two ResNet pathways to encode multi-scale information. The work of \cite{qiu2017learning} used 3D convolutional kernels for learning spatio-temporal features from video sequences.
%
%
Recently, graph-based methods proposed spatial-temporal graphs to model relationships between joints of the human body and produced state-of-the-art performance on popular action recognition datasets. These methods treat joints as nodes of the graph and bones as edges of the graph based on the pre-designed anatomy of human body. In this context, ST-GCN \cite{li2019spatio} proposed spatial graph convolutions with interleaving temporal convolutions to model relationships between skeleton joints for action recognition. The method of \cite{li2019actional} proposed AS-GCN which augmented spatial graph convolution with human poses for improved action recognition. The 2s-AGCN method of \cite{shi2019two} proposed graphs with self-attention mechanisms and used a two-stream ensemble with skeleton bone information to enhance action recognition accuracy. The method of \cite{shi2019skeleton} also used skeleton bone features, but instead of using an ensemble, their method jointly updates the joint and bone features through a spatial feature aggregation mechanism in the graph. The method of \cite{gao2019optimized} used a technique to fuse every three frames over the skeleton graph sequence and used cross-space-time edges between adjacent frames for temporal context.
In this work, we adopt a holistic approach to encode actions in terms of spatial and motion information extracted from face, hands, and body using visual and graph-based data representations. This approach has not been investigated before and it makes our work distinct from the existing studies which only consider one of the skeleton modalities individually. For instance methods like MS-G3d \cite{liu2020disentangling}, Shift-GCN \cite{cheng2020skeleton} and ST-GCN \cite{yan2018spatial} used only body skeleton data. Other methods like P-CNN \cite{cheron2015p}, P-I3D \cite{inproceedings}, and the methods of \cite{baradel2017pose,garcia2018first,lei2012fine,li2017action}, used body, hand or facial data alone. The method of \cite{liu2019skepxels} transformed only body joints spatial data into image representations. On the contrary, our Deep Action Stamps constitute graph-based and image-based data representations encoding spatial as well as motion information of body, hands, and face skeletons. This makes our data representation novel compared to the existing data representations.
Furthermore, our model presents a specially designed architecture to learn multi-modal convolutional and structural features from the spatial and motion data of body, face, and hands joints using a combined objective function. Experiments show that our model produces considerable improvements in terms of recognition accuracy and computational efficiency compared to the previous methods. It is also to be noted that the proposed ensemble architecture can be expanded with more advanced graph layers or temporal convolutions to further increase the model’s capability for learning spatial and temporal dependencies between joints (we left this for future work). These attributes make our model unique and open new possibilities for advancing action recognition from videos.
\subsection{The Proposed Deep Action Stamps}\label{deepacts}
\begin{figure*}[t!]
	\begin{center}
		\includegraphics[trim=0.1cm 0.2cm 0.1cm 3.5cm,clip,width=1.0\linewidth,keepaspectratio]{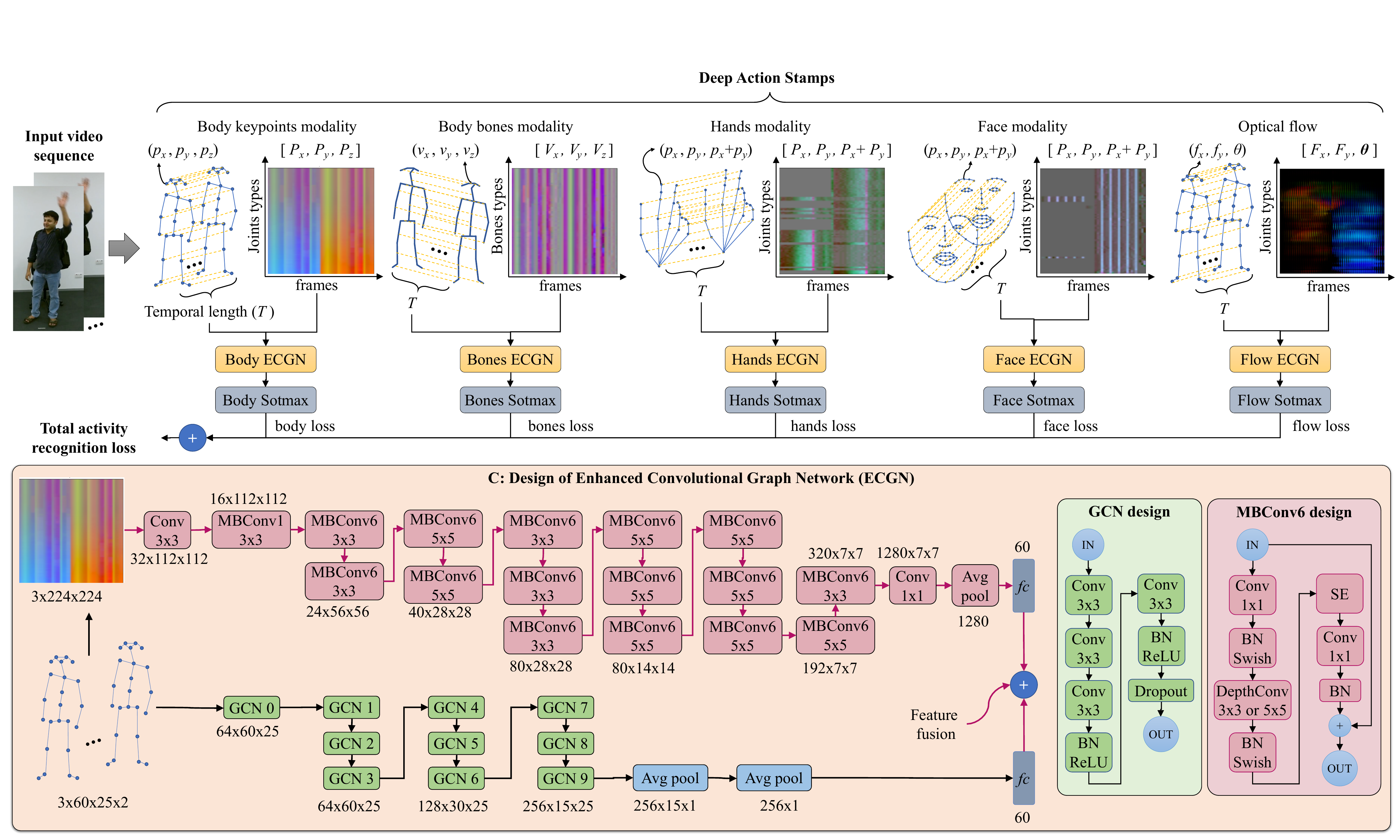}
		\vspace{-20pt}
		\caption{Overview of the proposed Deep Actions Stamps (DeepActs) and the DeepActsNet model. DeepActs encode actions from an RGB video in terms of pose and motion information of the human body, face, and hands using graph-based and image-based representations. DeepActsNet ensmebles modality-specific sub-networks termed Enhanced Convolutional Graph Networks (ECGN) that learn features from DeepActs using bottleneck and graph convolutional layers. DeepActsNet is trained jointly in an end-to-end manner using a combined loss to learn the spatial and temporal dependencies between facial joints, hand joints, and body joints for action recognition.}
		\label{fig_network}
	\end{center}
\end{figure*}
\begin{figure*}[t!]
	\begin{center}
		\includegraphics[trim=0.1cm 0.2cm 0.1cm 0.3cm,clip,width=1.0\linewidth,keepaspectratio]{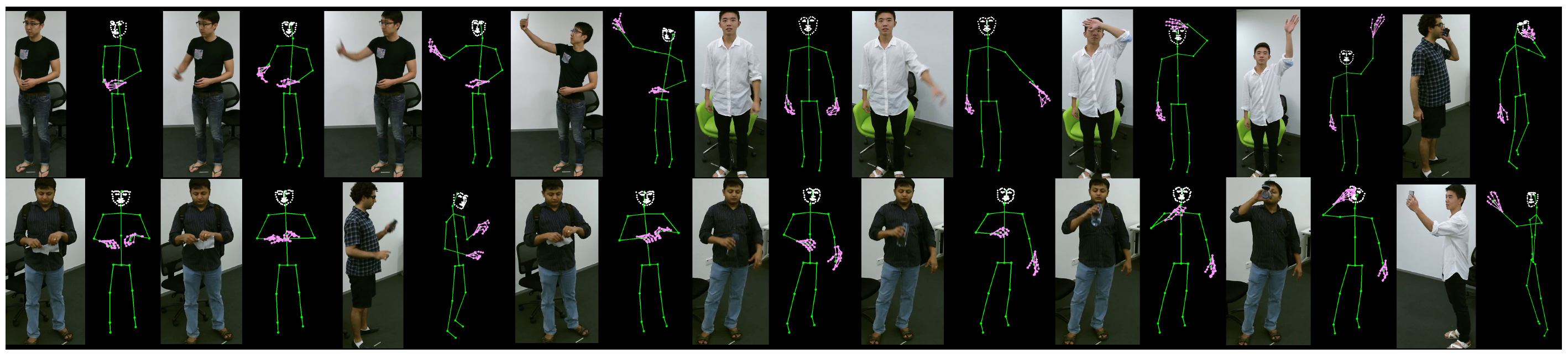}
		\vspace{-20pt}
		\caption{Visual representation of joints information extracted from human body, face, and hands for example video sequences.}
		\label{fig_pose}
	\end{center}
\end{figure*}
Deep Action Stamps (DeepActs) are composed of multi-modal graph-based and image-based data representations which encode actions in terms of spatial and motion information of body skeleton, skeletons of two hands, and face skeleton. 
Body skeleton data consists of 3D keypoints of 25 joints already provided by the authors of the NTU60 \cite{shahroudy2016ntu}, NTU120 \cite{liu2019ntu}, and SYSU \cite{hu2015jointly} datasets, and bones information which represent the difference of coordinates between two joints spatially connected as shown in Fig. \ref{fig_network}. Specifically, each bone is represented as a 3D vector pointing to its target joint form its source joint, encoding information both about the length and the direction of the vector. 
Hand skeleton data consists of 2D keypoints of 21 finger joints extracted from each hand. To extract hand skeleton data from videos, we trained our own Detectron2 \cite{wu2019detectron2} based hand pose estimator using the hands database of \cite{simon2017hand}. 
Face skeleton data consists of 2D keypoints information of 68 landmark positions as defined in \cite{sagonas2016300}. To extract face skeleton data from videos, we used a combination of RetinaFace \cite{deng2019retinaface} and the DLib facial landmark detector method of OpenCV. Fig. \ref{fig_pose} shows qualitative results of our multi-modal joints estimation for example videos of NTU60 dataset. Motion data consists of optical flow estimated between adjacent frames of a video sequence and represented with respect to the different joints types as shown in Fig. \ref{fig_network}.
Since, most of the joints information is estimated using deep learning based methods, we term the multi-modal data ``Deep Action Stamps".
Specifically, Deep Action Stamps contain five data modalities, where each modality is represented by a $C\times N_{f}\times N_{j}\times N_{p}-$dimensional graphical representation and a $C\times W\times H-$dimensional visual representation. 
Given sequential joints spatial and motion information from body, face, and hands, we construct DeepActs as follows. \textbf{First}, we construct a $C\times T\times N_{j}\times N_{p}-$dimensional graphical representation for each modality, where $C$ represents the number of feature channels, $T$ represents the temporal length (number of frames), $N_{j}$ represents the number of joints, and $N_{p}$ represents the number of people detected in the video sequence. 
For body skeleton, the feature channels ($C$) correspond to the $x$, $y$, and $z$ joint coordinates. For face and hand skeletons, the feature channels correspond to $x$, $y$, and $z=x+y$ values of 2D keypoints. For optical-flow, the feature channels represent the optical flow information in $x$ direction, $y$ direction, and the orientation component $\theta$.
\textbf{Next}, we construct visual representations for each data modality. For this, we reshape the $C\times T\times N_{j}\times N_{p}-$dimensional graphical representations to $C\times T\times N_{j}*N_{p}$ dimensions and normalize the channels between 0 and 255. We resize the visual representations to a fixed width and height ($C\times 224\times 224$) using bilinear interpolation. 
For a given joint $\boldsymbol{p}_{t,k}$ of type $k$ at frame $t$, the corresponding normalized pixel value is computed as:
\begin{equation}
\boldsymbol{d}=255\times \frac{(\boldsymbol{p}_{t,k}-c_{\min }) }{(c_{\max }-c_{\min }) },
\end{equation}
where $c_{min}$, $c_{max}$ correspond to the minimum and the maximum values of all the joint coordinates in the data respectively. Fig. \ref{fig_network} shows the graphical and visual representations of Deep Actions Stamps for an example video sequence.
\subsection{The Proposed DeepActsNet}
DeepActsNet is composed of modality-specific sub-networks termed Enhanced Convolutional Graph Networks (ECGN) connected in an ensemble architecture as shown in Fig. \ref{fig_network}. Each ECGN is composed of a convolutional branch and a graph branch which learn convolutional and structural features from Deep Action Stamps, respectively. The features are combined through summation and fed into a SoftMax operation to compute a modality-specific loss. The total loss is the sum of the loss of all data modalities. 
\subsubsection{ECGN Convolutional Branch}
It starts with a $3\times3$ convolution followed by 16 ``MBConv" modules. MBConv is an inverted bottleneck convolution module of \cite{tan2019efficientnet} which is composed of a $1\times 1$ convolution followed by Batch-Normalization, $3\times 3$ or $5\times 5$ depth-wise convolutions of \cite{sandler2018mobilenetv2}, a Squeeze-and-Excitation block (SE) of \cite{hu2018squeeze}, and a residual connection as shown in Fig. \ref{fig_network}. The network uses the Swish activation function.
The MBConv modules are followed by a $1\times1$ convolution, a global averaging operation and a fully connected layer $fc$ which learns probabilistic distributions of the features with respect to the target classes.
\subsubsection{ECGN Graph Branch}
It consists of a graph network with spatial and temporal connections which are defined through a fixed adjacency matrix to represent an action video. Fig. \ref{fig_network} shows examples of graphs defined on body, face, hands, and optical flow data, where the joints are represented as nodes and their spatial connections are represented as edges (the solid lines). For the temporal dimension, the corresponding joints between adjacent frames are connected through temporal edges (dotted yellow lines).
Consider an undirected graph at each time step $\mathcal{G}_{t}=(\mathcal{V}_{t},\mathcal{E}_{t})$, where $\mathcal{V}_{t}=\{v_{t1},...,v_{tN}\}$ is the set of $N$ nodes representing joints, and $\mathcal{E}_{t}=\{(v_{ti},v_{tj}):v_{ti},v_{tj}\in\mathcal{V}_{t},v_{ti}\sim v_{tj}\}$ is the set of edges in the graph, representing connections between the joints defined by an adjacency matrix $\mathbf{A}_{t}\in\mathbb{R}^{N\times N}$. $v_{ti}\sim v_{tj}$ represents that the node $i$ and node $j$ are connected with an undirected edge based on the anatomy shown in Fig. \ref{fig_network}. The adjacency matrix $A_{t}$ is defined as:
\begin{equation}
A_{t}(i,j)=\left\{\begin{matrix}
1 &if (v_{ti},v_{tj})\in \mathcal{E}_{t} \\ 
0 & otherwise
\end{matrix}\right.
\end{equation}
Given the graph defined above, multiple layers with graph convolution operations (GCN) are applied on the graph as shown in Fig. \ref{fig_network}. For an input tensor ${f}_{in}$, the output of a graph convolution can be written as:
\begin{equation}
\boldsymbol{f}_{out}=\sum_{k}^{K_{v}}W_{k}(\boldsymbol{f}_{in}A_{k})\circ M_{k},
\end{equation}
where $K_v$ denotes the kernel size of the spatial dimension, $A_k$ is the adjacency matrix. $W_{k}$ is the $C_{out}\times C_{in}\times 1 \times 1$ weight vector of the $1\times 1$ convolution operation. $\circ$ represents the dot product. The term $M_{k}\in \mathbb{R}^{N\times N}$ represents a learnable mask to increase the effectiveness of the convolution as used in \cite{yan2018spatial}. 
There are 10 GCN blocks in the graph branch. The output from the last GCN block is average pooled along both the temporal and joint dimensions resulting in a $256\times1-$dimensional tensor which is fed into a linear layer $fc$ to learn probabilistic distributions of the features with respect to the target classes. Each GCN block is composed of a set of four $3\times3$ convolution layers with Batch Normalization, ReLU, and a dropout as shown in Fig. \ref{fig_network}.
Note that the input to our GCN is always $C\times T\times N_{j}\times N_{p}$-dimensional graphical representation for each modality, where the data from multiple persons are
concatenated along the $N_{p}$ dimensions. For single person cases, the respective dimensions are filled with zeros.
\subsubsection{Training Loss}
Given $N$ training samples $X=\{x_{i}\}_{i=1}^N$ from $M$ classes, we denote the corresponding action class label set as $Y=\{y_{i}\}_{i=1}^M, y_{i}\in {1, ..., M}$. 
We also denote the ECGN classifiers in DeepActsNet by $\Theta=\{\theta_{body},\theta_{bones},\theta_{face},\theta_{hands},\theta_{flow}\}$. A softmax layer is applied after each ECGN classifier. It is given by:
\begin{equation}
q_{i}^{k}=\frac{exp(z_{i}^{k})}{\sum_{j}^k exp(z_{j}^k)},
\end{equation}
where $z$ represents the combined output of the convolutional and graph branch of the ECGN shown by feature fusion in Fig. \ref{fig_network}. The term $q_{i}^k\in\mathbb{R}^{\mathbb{M}}$ represents the $i^{th}$ class probability of the ECGN classifier $\theta_{k}$.
The loss function of the whole network is given by the sum of the loss of the modality-specific ECGN classifiers. It can be written as:
\begin{equation}
loss = \sum_{k\in\Theta}CE(q^{k},y),
\end{equation}
where $CE$ represents a CrossEntropy function.
\section{Experiments}\label{exp}
We conducted extensive experiments on three datasets which have been widely used in previous works for action recognition.
\subsection{Datasets}
\subsubsection{NTU60 RGB+D Dataset}
NTU60 \cite{shahroudy2016ntu} is a large-scale action recognition dataset consisting of 56,880 skeleton sequences categorized into 60 classes (comprising of daily, mutual (more than one person), and health-related activities). The skeleton sequences consist of 3D joint coordinates of 25 human body joints and their corresponding RGB videos. The data was collected by recording 40 distinct subjects (using a Microsoft Kinectv2 sensor) who performed the target activities at 17 different setup locations and under three different camera viewpoints [-45$^{\circ}$,0$^{\circ}$,45$^{\circ}$]. We followed the standard benchmark evaluation protocol as used in \cite{shahroudy2016ntu}: 1) Cross-subject (CS) setting, where half of the 40 subjects are included in training and the other half are used for testing, producing 40,091 and 16,487 training and testing examples respectively. 2) Cross-View (CV) setting, where all 18,932 samples captured from camera 1 are used for testing and the remaining 37,646 samples are used for training.
\subsubsection{NTU120 RGB+D Dataset}
NTU120 \cite{liu2019ntu} extends NTU60 dataset with an additional 57,367 skeleton sequences over 60 additional classes, yielding 114,480 activity samples over 120 action classes recorded from 106 distinct subjects and 32 camera viewpoints. For evaluation on this dataset, we followed the standard benchmark settings as used in \cite{liu2019ntu}. For the Cross Subject (CS) setting, half of 106 subjects are included in training and the rest are included in testing. For the Cross-Setup (CSet) setting, activity samples from even numbered setups are used for training and those from odd number setups are used for testing.
\subsubsection{SYSU 3D Human-Object Interaction Dataset}
SYSU \cite{hu2015jointly} consists of 480 activity samples collected by Kinect camera. For each activity sample, the RGB frames and skeleton data of 20 body joints are provided. The data was collected by recording 40 human subjects performing 12 different activities. 
We followed the standard benchmark settings as used in \cite{hu2015jointly}. For the Cross Subject (CS) setting, half of the subjects are used for training and the other half for testing. For the Same Subject (SS) setting, half of the samples for each activity are used in training and the other half for testing. We evaluate our models for 30-folds and report the mean accuracy for each setting.
\begin{figure*}[htbp]
	\begin{center}
		\includegraphics[trim=0.0cm 0.5cm 0.0cm 0.1cm,clip,width=1.0\linewidth,keepaspectratio]{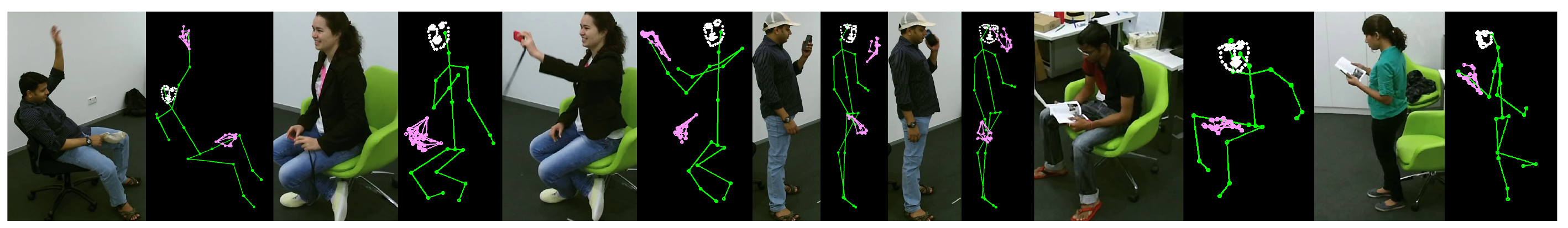}
		\vspace{-20pt}
		\caption{Examples where hand joints estimated by body pose estimator contain errors. On the other hand, finger keypoints estimated by our hand pose estimator produces correct joints locations.}
		\label{fig_errors}
	\end{center}
\end{figure*}
\subsection{Implementation Details}
We trained DeepActsNet using SGD with a momentum of 0.9, a batch size of 32, an initial learning rate of 1e-4 and an exponential LR decay with a factor of 0.1. Weight decay was set to 0.0005. For NTU60 and NTU120 datasets, all joints data was transformed to a fixed temporal length of $T=60$ frames. For SYSU dataset, the data was transformed to $T=200$ frames. No data augmentation was used.
For NTU-60 and NTU-120 datasets, we followed the data pre-processing of \cite{shahroudy2016ntu,zhang2019view}, where we removed falsely detected bodies (corresponding to background objects in the scene) through denoising based on frame length and spread of the joint locations along $x$ and $y$ axes for each body. We removed the bodies for which the frame length was less than 11 or where the $x$ spread was more than 0.8 of the $y$ spread of joint locations.
\subsection{Results}
\begin{table*}[htbp]
	\caption{Ablation study of Deep Action Stamps and the proposed DeepActsNet on the NTU60 \cite{shahroudy2016ntu}, NTU120 \cite{liu2019ntu}, and SYSU \cite{hu2015jointly} datasets.}
	\vspace{-10pt}
	\label{table_ablation}
	\centering
	\setlength\tabcolsep{4.0pt}
	\begin{tabular}{@{}lcccccccccccc@{}}
		\toprule
		&\multirow{2}{*}{Model}&\multirow{2}{*}{Modality}&\multicolumn{2}{c}{NTU60}&\multicolumn{2}{c}{NTU120}&\multicolumn{2}{c}{SYSU} &Param.&\multirow{2}{*}{FLOPS}&Time\\
		&&&CS(\%)& CV(\%)& CS(\%) & CSet(\%)& CS(\%) & SS(\%)&million&&ms\\
		\midrule
		A) & \multirow{5}{*}{DeepActsNet} 	 & Face joints & 44.5 & 45.6 & 31.6 & 33.0 & 53.8 & 41.6 & 63.9 & 0.16 & 50.6 \\
		B) & 	 & Hands joints & 52.1 & 52.9 & 36.4 & 37.0 & 76.1 & 60.8  \\
		C) & 	 & Body bones & 83.7 & 88.6 & 77.7 & 79.1 & 83.3 & 82.0  \\
		D) & Conv-stream 	& Body joints & 85.4 & 90.1 & 78.0 & 79.4 & 81.9 & 81.8  \\
		E) & 	 & Optical flow & 83.9 & 90.4 & 73.9 & 76.4 & 87.5 & 74.8  \\
		F) & 	 & All & 92.1 & 96.4 & 86.9 & 88.8 & 92.8 & 88.5 & 319.7 & 0.80 & 236 \\
		\hline
		G) & \multirow{5}{*}{DeepActsNet} 	 & Face joints & 46.9 & 47.0 & 37.0 & 38.1 & 42.8 & 37.0 & 3.1 & 6.4 & 8 \\
		H) & 	 & Hands joints & 51.9 & 55.3 & 39.9 & 41.3 & 69.4 & 61.3  \\
		I) & 	 & Body bones & 84.4 & 85.9 & 78.8 & 79.1 & 81.3 & 68.3  \\
		J) & Graph-stream 	 & Body joints & 85.5 & 89.7 & 78.3 & 80.7 & 85.6 & 74.3  \\
		K) & 	 & Optical flow & 83.2 & 89.7 & 72.2 & 73.7 & 82.1 & 70.2  \\
		L) & 	 & All & 92.6 & 95.6 & 87.7 & 89.4 & 91.4 & 84.9 & 15.5&32.6&40 \\
		\hline
		M) & \multirow{5}{*}{DeepActsNet} 	 & Body & 89.4 & 93.5 & 83.5 & 85.3 & 88.5 & 85.2 &67.0&6.7&53 \\
		N) & 	 & Body+Face & 88.0 & 91.8 & 82.6 & 83.7 & 86.4 & 82.2 &134&13.4&		108 \\
		O) & 	 & Body+Hands & 89.8 & 93.8 & 84.0 & 85.9 & 90.7 & 87.7 &134&13.4&		108 \\
		P) & Conv+Graph 	 & Body+Flow & 93.4 & 97.5 & 88.5 & 90.5 & 93.5 & 88.7 &134&13.4&		108 \\
		Q) & 	 & Body+Face+Hands & 90.2 & 93.7 & 84.3 & 85.9 & 90.2 & 86.5 &201&20.0&161 \\
		R) & 	 & Body+Face+Hands+Flow & 93.4 & 97.2 & 88.2 & 90.1 & 93.9 & 89.6 &268&26.8&215 \\
		S) & 	 & All &\textbf{94.3} & \textbf{97.6} & \textbf{90.2} & \textbf{91.8} & \textbf{93.9} & \textbf{90.1} &335&33.4&268 \\
		\hline
		T) & DeepActsNet-lite 	 & All & 94.2 & 97.3 & 89.7 & 91.6 & 93.4 & 88.9 &258&7.1&206 \\
		U) & DeepActsNet-tiny 	 & All & 94.0 & 97.4 & 89.3 & 91.3 & 93.7 & 88.4 &19.4&6.6&62 \\
		\bottomrule
	\end{tabular}
\end{table*}
\subsubsection{Ablation Study of Deep Action Stamps}
Here we evaluate the significance of the feature channels of DeepActs for action recognition. For this, we investigate three settings of DeepActsNet: Conv-stream, which only ensembles the convolutional branches of ECGN; Graph-stream, which only ensembles the graph branches of ECGN, and Conv+Graph stream which ensembles both the convolutional and graph branches of ECGN for all data modalities.
Table \ref{table_ablation} shows that the recognition accuracy consistently increase on all the datasets as we combine more feature channels of DeepActs to train the models. 
For instance, the Conv-stream using all feature channels produced improvements of 7.6\% and 4.2\% in the cross-subject and cross-view recognition accuracy on the NTU60 dataset compared to the models trained using body joints information alone (see row-D in Table \ref{table_ablation} and row-F in Table \ref{table_ablation}). 
Furthermore, the Graph-stream improves the cross-subject accuracy from 84.9\% to 93.3\% and cross-view accuracy from 91.9\% to 97.2\% when all features channels were used compared to the case of using body joints information alone (see row-J in Table \ref{table_ablation} and row-L in Table \ref{table_ablation}). 
\\
\indent
Table \ref{table_ablation} also shows the benefits of using multi-modal DeepActs for the Conv+Graph models. For instance, the addition of facial and hands data produced improvements of 4\% and 3.9\% in the cross-subject and cross-setup recognition accuracy on the NTU120 dataset as shown in row-O of Table \ref{table_ablation}. The addition of motion information to the facial and hands data yields further improvements of around 2\% and 2.1\% in the cross-subject and cross-setup recognition accuracy as shown in row-P of Table \ref{table_ablation}. The addition of bones information produces further improvements of upto 1\% and 1.4\% in the cross-subject and cross-setup accuracy, respectively as shown in row-Q of Table \ref{table_ablation}.
\begin{table*}[htbp]
	\caption{Improvements in class-wise accuracy of DeepActsNet for the 60 classes of the NTU60 dataset \cite{shahroudy2016ntu} on the Cross View (CV) and Cross Subject (CS) settings, using body+hands data (Model A), using body+flow data (Model B), using body+face+hands (Model C), and using body+face+hands+bones+flow (Model D). We used the body joints data as the performance baselines (values in black). Values in green and red are accuracy differences compared to the baselines.}
	\vspace{-10pt}
	\label{table1}
	\centering
	\setlength\tabcolsep{4.0pt}
	\begin{tabular}{@{}clcccccccccc@{}}
		\toprule   
		&&Baselines&	{Model A} &Model B&Model C&Model D&Baselines&Model A&Model B& Model C&Model D\\			
		&{Action Class}&CV (\%) &CV (\%)&{CV (\%)} &CV (\%)&CV (\%)&CS (\%)&CS(\%)&CS (\%)&CS (\%)&CS (\%)\\					
		\midrule
		1 & drink water                              	 & 94.9 	 & \textcolor{red}{  -2.2 } 	 & \textcolor{ForestGreen}{ + 3.8 } 	 & \textcolor{ForestGreen}{ + 0.9 }  & \textcolor{ForestGreen}{ + 3.3 } 	 & 84.7 	 & \textcolor{red}{  -4.0 } 	 & \textcolor{ForestGreen}{ + 10.2 } 	 & \textcolor{ForestGreen}{ + 1.5 }	 & \textcolor{ForestGreen}{ + 9.7 }\\
		2 & eat meal/snack                           	 & 82.3 	 & \textcolor{ForestGreen}{ + 3.8 } 	 & \textcolor{ForestGreen}{ + 13.6 } 	 & \textcolor{ForestGreen}{ + 0.6 } 	 & \textcolor{ForestGreen}{ + 14.7 } 	 & 72.7 	 & \textcolor{ForestGreen}{ + 2.2 } 	 & \textcolor{ForestGreen}{ + 8.4 } 	 & \textcolor{ForestGreen}{ + 6.2 }	 & \textcolor{ForestGreen}{ + 14.5 }\\
		3 & brushing teeth                           	 & 92.7 	 & \textcolor{red}{  -1.9 } 	 & \textcolor{ForestGreen}{ + 6.3 } 	 & \textcolor{ForestGreen}{ + 0.6 }  & \textcolor{ForestGreen}{ + 4.2 } 	 & 83.5 	 & \textcolor{ForestGreen}{ + 5.9 } 	 & \textcolor{ForestGreen}{ + 10.6 } 	 & \textcolor{ForestGreen}{ + 3.7 }	 & \textcolor{ForestGreen}{ + 8.8 }\\
		4 & brushing hair                            	 & 95.3 	 & \textcolor{ForestGreen}{ + 0.6 } 	 & \textcolor{ForestGreen}{ + 2.5 } 	 & \textcolor{ForestGreen}{ + 1.3 } 	 & \textcolor{ForestGreen}{ + 2.7 } 	 & 87.9 	 & \textcolor{ForestGreen}{ + 0.4 } 	 & \textcolor{ForestGreen}{ + 3.3 } 	 & \textcolor{red}{  -3.3 }	 & \textcolor{ForestGreen}{ + 6.6 }\\
		5 & drop                                     	 & 96.5 	 & \textcolor{ForestGreen}{ + 0.3 } 	 & \textcolor{ForestGreen}{ + 3.5 } 	 & \textcolor{ForestGreen}{ + 0.9 } 	 & \textcolor{ForestGreen}{ + 2.9 } 	 & 89.8 	 & \textcolor{red}{  -10.5 } 	 & \textcolor{ForestGreen}{ + 7.3 } 	 & \textcolor{red}{  -6.9 }	 & \textcolor{ForestGreen}{ + 7.6 }\\
		6 & pickup                                   	 & 94.3 	 & \textcolor{red}{  -0.9 } 	 & \textcolor{ForestGreen}{ + 5.1 } 	 & \textcolor{ForestGreen}{ + 3.2 }  & \textcolor{ForestGreen}{ + 5.2 } 	 & 97.1 	 & \textcolor{red}{  -3.3 } 	 & \textcolor{ForestGreen}{ + 0.7 } 	 & \textcolor{red}{  -1.1 }	 & \textcolor{ForestGreen}{ + 0.7 }\\
		7 & throw                                    	 & 98.7 	 & \textcolor{ForestGreen}{ + 0.0 } 	 & \textcolor{ForestGreen}{ + 0.9 } 	 & \textcolor{red}{  -0.3 }  & \textcolor{ForestGreen}{ + 0.8 } 	 & 95.6 	 & \textcolor{red}{  -1.8 } 	 & \textcolor{red}{  -0.7 } 	 & \textcolor{red}{  -0.4 }	 & \textcolor{ForestGreen}{ + 2.2 }\\
		8 & sitting down                             	 & 97.8 	 & \textcolor{ForestGreen}{ + 1.3 } 	 & \textcolor{ForestGreen}{ + 1.6 } 	 & \textcolor{ForestGreen}{ + 2.2 } 	 & \textcolor{ForestGreen}{ + 2.2 } 	 & 96.3 	 & \textcolor{ForestGreen}{ + 0.4 } 	 & \textcolor{ForestGreen}{ + 1.1 } 	 & \textcolor{ForestGreen}{ + 1.8 }	 & \textcolor{ForestGreen}{ + 2.7 }\\
		9 & standing up (from sitting position)      	 & 98.7 	 & \textcolor{ForestGreen}{ + 0.6 } 	 & \textcolor{ForestGreen}{ + 0.9 } 	 & \textcolor{ForestGreen}{ + 0.6 } 	 & \textcolor{ForestGreen}{ + 0.9 } 	 & 98.5 	 & \textcolor{ForestGreen}{ + 0.0 } 	 & \textcolor{ForestGreen}{ + 0.4 } 	 & \textcolor{ForestGreen}{ + 1.1 }	 & \textcolor{ForestGreen}{ + 0.7 }\\
		10 & clapping                                 	 & 93.0 	 & \textcolor{ForestGreen}{ + 1.6 } 	 & \textcolor{ForestGreen}{ + 3.5 } 	 & \textcolor{red}{  -0.6 }  & \textcolor{ForestGreen}{ + 1.3 } 	 & 71.4 	 & \textcolor{ForestGreen}{ + 14.3 } 	 & \textcolor{ForestGreen}{ + 13.2 } 	 & \textcolor{ForestGreen}{ + 16.1 } & \textcolor{ForestGreen}{ + 21.0 }\\
		11 & reading                                  	 & 85.7 	 & \textcolor{red}{  -2.9 } 	 & \textcolor{ForestGreen}{ + 6.3 } 	 & \textcolor{red}{  -1.0 } 	 & \textcolor{ForestGreen}{ + 3.6 } 	 & 56.4 	 & \textcolor{ForestGreen}{ + 0.7 } 	 & \textcolor{ForestGreen}{ + 13.9 } 	 & \textcolor{red}{  -1.5 }	 & \textcolor{ForestGreen}{ + 20.0 }\\
		12 & writing                                  	 & 57.8 	 & \textcolor{ForestGreen}{ + 10.2 } 	 & \textcolor{ForestGreen}{ + 19.0 } 	 & \textcolor{ForestGreen}{ + 7.3 } 	 & \textcolor{ForestGreen}{ + 22.3 } 	 & 52.2 	 & \textcolor{red}{  -0.4 } 	 & \textcolor{ForestGreen}{ + 11.4 } 	 & \textcolor{ForestGreen}{ + 10.3 } & \textcolor{ForestGreen}{ + 18.5 }\\
		13 & tear up paper                            	 & 94.6 	 & \textcolor{ForestGreen}{ + 1.6 } 	 & \textcolor{ForestGreen}{ + 2.8 } 	 & \textcolor{ForestGreen}{ + 0.3 } 	 & \textcolor{ForestGreen}{ + 4.3 } 	 & 92.6 	 & \textcolor{ForestGreen}{ + 2.2 } 	 & \textcolor{ForestGreen}{ + 3.7 } 	 & \textcolor{red}{  -0.7 }	 & \textcolor{ForestGreen}{ + 3.9 }\\
		14 & wear jacket                              	 & 99.7 	 & \textcolor{ForestGreen}{ + 0.3 } 	 & \textcolor{ForestGreen}{ + 0.0 } 	 & \textcolor{ForestGreen}{ + 0.3 } 	 & \textcolor{ForestGreen}{ + 0.3 } 	 & 98.5 	 & \textcolor{ForestGreen}{ + 0.4 } 	 & \textcolor{ForestGreen}{ + 0.4 } 	 & \textcolor{ForestGreen}{ + 0.7 }	 & \textcolor{red}{  -0.5 }\\
		15 & take off jacket                          	 & 99.1 	 & \textcolor{red}{  -0.3 } 	 & \textcolor{ForestGreen}{ + 0.9 } 	 & \textcolor{red}{  -0.9 } 	 & \textcolor{ForestGreen}{ + 0.9 } 	 & 96.4 	 & \textcolor{red}{  -0.7 } 	 & \textcolor{ForestGreen}{ + 2.2 } 	 & \textcolor{red}{  -1.1 }	 & \textcolor{ForestGreen}{ + 1.3 }\\
		16 & wear a shoe                              	 & 93.3 	 & \textcolor{red}{  -0.3 } 	 & \textcolor{ForestGreen}{ + 5.1 } 	 & \textcolor{ForestGreen}{ + 1.9 }  & \textcolor{ForestGreen}{ + 2.8 } 	 & 78.8 	 & \textcolor{ForestGreen}{ + 5.1 } 	 & \textcolor{ForestGreen}{ + 14.3 } 	 & \textcolor{ForestGreen}{ + 4.0 }	 & \textcolor{ForestGreen}{ + 12.7 }\\
		17 & take off a shoe                          	 & 91.1 	 & \textcolor{ForestGreen}{ + 0.0 } 	 & \textcolor{ForestGreen}{ + 5.7 } 	 & \textcolor{red}{  -4.1 }  & \textcolor{ForestGreen}{ + 5.3 } 	 & 78.5 	 & \textcolor{red}{  -5.5 } 	 & \textcolor{ForestGreen}{ + 8.0 } 	 & \textcolor{red}{  -3.3 }	 & \textcolor{ForestGreen}{ + 10.8 }\\
		18 & wear on glasses                          	 & 92.7 	 & \textcolor{ForestGreen}{ + 0.3 } 	 & \textcolor{ForestGreen}{ + 4.7 } 	 & \textcolor{ForestGreen}{ + 0.6 } 	 & \textcolor{ForestGreen}{ + 5.5 } 	 & 89.7 	 & \textcolor{ForestGreen}{ + 3.7 } 	 & \textcolor{ForestGreen}{ + 3.3 } 	 & \textcolor{ForestGreen}{ + 4.0 }	 & \textcolor{ForestGreen}{ + 6.0 }\\
		19 & take off glasses                         	 & 96.2 	 & \textcolor{ForestGreen}{ + 0.6 } 	 & \textcolor{ForestGreen}{ + 3.5 } 	 & \textcolor{ForestGreen}{ + 0.6 } 	 & \textcolor{ForestGreen}{ + 2.9 } 	 & 96.7 	 & \textcolor{red}{  -1.1 } 	 & \textcolor{ForestGreen}{ + 2.6 } 	 & \textcolor{red}{  -2.6 }	 & \textcolor{red}{  -1.0 }\\
		20 & put on a hat/cap                         	 & 98.7 	 & \textcolor{ForestGreen}{ + 0.6 } 	 & \textcolor{ForestGreen}{ + 1.0 } 	 & \textcolor{ForestGreen}{ + 1.0 } 	 & \textcolor{ForestGreen}{ + 0.6 } 	 & 96.7 	 & \textcolor{red}{  -0.7 } 	 & \textcolor{ForestGreen}{ + 1.5 } 	 & \textcolor{red}{  -0.4 }	 & \textcolor{ForestGreen}{ + 2.2 }\\
		21 & take off a hat/cap                       	 & 99.4 	 & \textcolor{red}{  -0.3 } 	 & \textcolor{ForestGreen}{ + 0.3 } 	 & \textcolor{red}{  -0.3 } 	 & \textcolor{ForestGreen}{ + 0.5 } 	 & 98.2 	 & \textcolor{red}{  -0.4 } 	 & \textcolor{red}{  -0.4 } 	 & \textcolor{ForestGreen}{ + 0.7 }	 & \textcolor{ForestGreen}{ + 1.3 }\\
		22 & cheer up                                 	 & 98.1 	 & \textcolor{ForestGreen}{ + 0.3 } 	 & \textcolor{ForestGreen}{ + 1.9 } 	 & \textcolor{ForestGreen}{ + 1.3 } 	 & \textcolor{ForestGreen}{ + 1.6 } 	 & 93.1 	 & \textcolor{ForestGreen}{ + 1.1 } 	 & \textcolor{ForestGreen}{ + 0.4 } 	 & \textcolor{ForestGreen}{ + 2.6 }	 & \textcolor{ForestGreen}{ + 4.4 }\\
		23 & hand waving                              	 & 96.5 	 & \textcolor{red}{  -1.3 } 	 & \textcolor{ForestGreen}{ + 1.3 } 	 & \textcolor{red}{  -4.1 } 	 & \textcolor{ForestGreen}{ + 2.0 } 	 & 92.0 	 & \textcolor{ForestGreen}{ + 1.5 } 	 & \textcolor{ForestGreen}{ + 1.8 } 	 & \textcolor{ForestGreen}{ + 0.0 }	 & \textcolor{ForestGreen}{ + 2.7 }\\
		24 & kicking something                        	 & 98.1 	 & \textcolor{red}{  -2.8 } 	 & \textcolor{ForestGreen}{ + 1.3 } 	 & \textcolor{red}{  -1.9 } 	 & \textcolor{ForestGreen}{ + 1.0 } 	 & 98.6 	 & \textcolor{red}{  -5.8 } 	 & \textcolor{red}{  -1.4 } 	 & \textcolor{red}{  -4.7 }	 & \textcolor{red}{  -2.5 }\\
		25 & reach into pocket                        	 & 93.0 	 & \textcolor{red}{  -1.6 } 	 & \textcolor{ForestGreen}{ + 4.4 } 	 & \textcolor{red}{  -0.3 } 	 & \textcolor{ForestGreen}{ + 5.9 } 	 & 85.4 	 & \textcolor{red}{  -0.4 } 	 & \textcolor{ForestGreen}{ + 3.6 } 	 & \textcolor{red}{  -0.4 }	 & \textcolor{ForestGreen}{ + 6.5 }\\
		26 & hopping (one foot jumping)               	 & 99.4 	 & \textcolor{ForestGreen}{ + 0.3 } 	 & \textcolor{ForestGreen}{ + 0.6 } 	 & \textcolor{ForestGreen}{ + 0.6 } 	 & \textcolor{ForestGreen}{ + 0.6 } 	 & 98.9 	 & \textcolor{red}{  -2.2 } 	 & \textcolor{ForestGreen}{ + 0.0 } 	 & \textcolor{red}{  -0.4 }	 & \textcolor{ForestGreen}{ + 0.4 }\\
		27 & jump up                                  	 & 99.4 	 & \textcolor{ForestGreen}{ + 0.0 } 	 & \textcolor{ForestGreen}{ + 0.6 } 	 & \textcolor{ForestGreen}{ + 0.6 } 	 & \textcolor{ForestGreen}{ + 0.6 } 	 & 100.0 	 & \textcolor{red}{  -4.7 } 	 & \textcolor{ForestGreen}{ + 0.0 } 	 & \textcolor{ForestGreen}{ + 0.0 }	 & \textcolor{red}{  -0.5 }\\
		28 & make a phone call/answer phone           	 & 93.7 	 & \textcolor{ForestGreen}{ + 1.6 } 	 & \textcolor{ForestGreen}{ + 5.4 } 	 & \textcolor{ForestGreen}{ + 1.3 } 	 & \textcolor{ForestGreen}{ + 4.4 } 	 & 89.1 	 & \textcolor{ForestGreen}{ + 2.5 } 	 & \textcolor{ForestGreen}{ + 6.9 } 	 & \textcolor{ForestGreen}{ + 3.3 }	 & \textcolor{ForestGreen}{ + 4.6 }\\
		29 & playing with phone/tablet                	 & 60.1 	 & \textcolor{ForestGreen}{ + 4.7 } 	 & \textcolor{ForestGreen}{ + 24.4 } 	 & \textcolor{red}{  -0.9 }  & \textcolor{ForestGreen}{ + 27.0 } 	 & 66.5 	 & \textcolor{ForestGreen}{ + 6.9 } 	 & \textcolor{ForestGreen}{ + 12.4 } 	 & \textcolor{ForestGreen}{ + 6.9 }	 & \textcolor{ForestGreen}{ + 11.8 }\\
		30 & typing on a keyboard                     	 & 75.9 	 & \textcolor{ForestGreen}{ + 12.0 } 	 & \textcolor{ForestGreen}{ + 19.0 } 	 & \textcolor{ForestGreen}{ + 10.1 } 	 & \textcolor{ForestGreen}{ + 11.4 } 	 & 72.0 	 & \textcolor{ForestGreen}{ + 20.0 } 	 & \textcolor{ForestGreen}{ + 13.1 } 	 & \textcolor{ForestGreen}{ + 12.0 }	 & \textcolor{ForestGreen}{ + 11.0 }\\
		31 & pointing to something with finger        	 & 91.7 	 & \textcolor{ForestGreen}{ + 1.3 } 	 & \textcolor{ForestGreen}{ + 2.9 } 	 & \textcolor{red}{  -1.0 }  & \textcolor{ForestGreen}{ + 4.2 } 	 & 77.5 	 & \textcolor{ForestGreen}{ + 6.2 } 	 & \textcolor{ForestGreen}{ + 1.4 } 	 & \textcolor{ForestGreen}{ + 6.9 }	 & \textcolor{ForestGreen}{ + 13.8 }\\
		32 & taking a selfie                          	 & 94.9 	 & \textcolor{ForestGreen}{ + 2.8 } 	 & \textcolor{ForestGreen}{ + 2.2 } 	 & \textcolor{ForestGreen}{ + 0.9 } 	 & \textcolor{ForestGreen}{ + 2.4 } 	 & 94.2 	 & \textcolor{ForestGreen}{ + 0.4 } 	 & \textcolor{ForestGreen}{ + 0.0 } 	 & \textcolor{red}{  -1.8 }	 & \textcolor{red}{  -2.0 }\\
		33 & check time (from watch)                  	 & 94.6 	 & \textcolor{ForestGreen}{ + 3.5 } 	 & \textcolor{ForestGreen}{ + 2.8 } 	 & \textcolor{ForestGreen}{ + 1.9 } 	 & \textcolor{ForestGreen}{ + 4.3 } 	 & 88.0 	 & \textcolor{ForestGreen}{ + 1.4 } 	 & \textcolor{ForestGreen}{ + 5.1 } 	 & \textcolor{ForestGreen}{ + 1.4 }	 & \textcolor{ForestGreen}{ + 8.5 }\\
		34 & rub two hands together                   	 & 79.4 	 & \textcolor{ForestGreen}{ + 5.4 } 	 & \textcolor{ForestGreen}{ + 14.9 } 	 & \textcolor{ForestGreen}{ + 3.5 } 	 & \textcolor{ForestGreen}{ + 14.5 } 	 & 86.2 	 & \textcolor{red}{  -0.4 } 	 & \textcolor{ForestGreen}{ + 7.2 } 	 & \textcolor{red}{  -2.5 }	 & \textcolor{ForestGreen}{ + 5.7 }\\
		35 & nod head/bow                             	 & 98.1 	 & \textcolor{red}{  -1.3 } 	 & \textcolor{ForestGreen}{ + 0.9 } 	 & \textcolor{red}{  -1.6 } 	 & \textcolor{ForestGreen}{ + 1.6 } 	 & 97.8 	 & \textcolor{red}{  -1.8 } 	 & \textcolor{ForestGreen}{ + 1.1 } 	 & \textcolor{red}{  -3.3 }	 & \textcolor{red}{  -1.1 }\\
		36 & shake head                               	 & 95.9 	 & \textcolor{ForestGreen}{ + 0.0 } 	 & \textcolor{ForestGreen}{ + 3.8 } 	 & \textcolor{ForestGreen}{ + 3.8 } 	 & \textcolor{ForestGreen}{ + 3.0 } 	 & 94.9 	 & \textcolor{ForestGreen}{ + 1.8 } 	 & \textcolor{ForestGreen}{ + 4.0 } 	 & \textcolor{ForestGreen}{ + 4.4 }	 & \textcolor{ForestGreen}{ + 4.7 }\\
		37 & wipe face                                	 & 90.8 	 & \textcolor{ForestGreen}{ + 0.6 } 	 & \textcolor{ForestGreen}{ + 7.9 } 	 & \textcolor{ForestGreen}{ + 4.4 } 	 & \textcolor{ForestGreen}{ + 7.8 } 	 & 89.5 	 & \textcolor{ForestGreen}{ + 4.0 } 	 & \textcolor{ForestGreen}{ + 6.2 } 	 & \textcolor{ForestGreen}{ + 2.5 }	 & \textcolor{ForestGreen}{ + 4.4 }\\
		38 & salute                                   	 & 98.4 	 & \textcolor{red}{  -2.8 } 	 & \textcolor{ForestGreen}{ + 0.3 } 	 & \textcolor{red}{  -5.7 } 	 & \textcolor{red}{  -0.3 } 	 & 92.4 	 & \textcolor{ForestGreen}{ + 0.4 } 	 & \textcolor{ForestGreen}{ + 4.0 } 	 & \textcolor{ForestGreen}{ + 0.0 }	 & \textcolor{ForestGreen}{ + 3.6 }\\
		39 & put the palms together                   	 & 95.6 	 & \textcolor{red}{  -1.6 } 	 & \textcolor{ForestGreen}{ + 1.9 } 	 & \textcolor{red}{  -4.7 } 	 & \textcolor{ForestGreen}{ + 3.0 } 	 & 96.4 	 & \textcolor{ForestGreen}{ + 0.7 } 	 & \textcolor{red}{  -1.1 } 	 & \textcolor{red}{  -3.3 }	 & \textcolor{ForestGreen}{ + 0.2 }\\
		40 & cross hands in front (say stop)          	 & 93.9 	 & \textcolor{ForestGreen}{ + 2.9 } 	 & \textcolor{ForestGreen}{ + 2.6 } 	 & \textcolor{ForestGreen}{ + 1.9 } 	 & \textcolor{ForestGreen}{ + 4.8 } 	 & 93.5 	 & \textcolor{ForestGreen}{ + 2.5 } 	 & \textcolor{ForestGreen}{ + 2.2 } 	 & \textcolor{ForestGreen}{ + 2.2 }	 & \textcolor{ForestGreen}{ + 4.3 }\\
		41 & sneeze/cough                             	 & 92.1 	 & \textcolor{ForestGreen}{ + 0.3 } 	 & \textcolor{ForestGreen}{ + 6.0 } 	 & \textcolor{red}{  -2.5 }  & \textcolor{ForestGreen}{ + 3.0 } 	 & 72.1 	 & \textcolor{red}{  -6.9 } 	 & \textcolor{ForestGreen}{ + 9.1 } 	 & \textcolor{ForestGreen}{ + 2.5 }	 & \textcolor{ForestGreen}{ + 15.7 }\\
		42 & staggering                               	 & 99.4 	 & \textcolor{ForestGreen}{ + 0.0 } 	 & \textcolor{ForestGreen}{ + 0.0 } 	 & \textcolor{ForestGreen}{ + 0.3 } 	 & \textcolor{ForestGreen}{ + 0.6 } 	 & 97.8 	 & \textcolor{ForestGreen}{ + 0.7 } 	 & \textcolor{ForestGreen}{ + 1.4 } 	 & \textcolor{ForestGreen}{ + 1.1 }	 & \textcolor{red}{  -1.0 }\\
		43 & falling                                  	 & 99.7 	 & \textcolor{ForestGreen}{ + 0.0 } 	 & \textcolor{ForestGreen}{ + 0.3 } 	 & \textcolor{ForestGreen}{ + 0.3 } 	 & \textcolor{ForestGreen}{ + 0.3 } 	 & 100.0 	 & \textcolor{red}{  -0.7 } 	 & \textcolor{red}{  -0.7 } 	 & \textcolor{red}{  -0.7 }	 & \textcolor{red}{  -0.9 }\\
		44 & touch head (headache)                    	 & 82.0 	 & \textcolor{red}{  -6.0 } 	 & \textcolor{ForestGreen}{ + 12.0 } 	 & \textcolor{red}{  -0.9 } 	 & \textcolor{ForestGreen}{ + 13.3 } 	 & 83.3 	 & \textcolor{red}{  -1.1 } 	 & \textcolor{ForestGreen}{ + 6.9 } 	 & \textcolor{ForestGreen}{ + 1.1 }	 & \textcolor{ForestGreen}{ + 1.1 }\\
		45 & touch chest (stomachache/heart pain)     	 & 92.1 	 & \textcolor{red}{  -3.2 } 	 & \textcolor{ForestGreen}{ + 2.5 } 	 & \textcolor{red}{  -5.4 } 	 & \textcolor{ForestGreen}{ + 3.9 } 	 & 91.3 	 & \textcolor{red}{  -7.2 } 	 & \textcolor{ForestGreen}{ + 0.4 } 	 & \textcolor{red}{  -9.4 }	 & \textcolor{ForestGreen}{ + 2.7 }\\
		46 & touch back (backache)                    	 & 88.9 	 & \textcolor{red}{  -3.2 } 	 & \textcolor{ForestGreen}{ + 7.9 } 	 & \textcolor{red}{  -2.5 } 	 & \textcolor{ForestGreen}{ + 8.5 } 	 & 91.3 	 & \textcolor{ForestGreen}{ + 0.0 } 	 & \textcolor{ForestGreen}{ + 8.0 } 	 & \textcolor{red}{  -1.4 }	 & \textcolor{ForestGreen}{ + 4.8 }\\
		47 & touch neck (neckache)                    	 & 96.5 	 & \textcolor{red}{  -0.6 } 	 & \textcolor{ForestGreen}{ + 1.6 } 	 & \textcolor{red}{  -2.8 } 	 & \textcolor{ForestGreen}{ + 1.3 } 	 & 80.8 	 & \textcolor{ForestGreen}{ + 2.5 } 	 & \textcolor{ForestGreen}{ + 11.6 } 	 & \textcolor{ForestGreen}{ + 1.1 }	 & \textcolor{ForestGreen}{ + 13.6 }\\
		48 & nausea or vomiting condition             	 & 90.5 	 & \textcolor{ForestGreen}{ + 0.6 } 	 & \textcolor{ForestGreen}{ + 5.1 } 	 & \textcolor{ForestGreen}{ + 3.8 } 	 & \textcolor{ForestGreen}{ + 5.2 } 	 & 89.1 	 & \textcolor{ForestGreen}{ + 0.4 } 	 & \textcolor{ForestGreen}{ + 2.5 } 	 & \textcolor{ForestGreen}{ + 4.0 }	 & \textcolor{ForestGreen}{ + 2.6 }\\
		49 & use a fan (with hand or paper)           	 & 96.8 	 & \textcolor{ForestGreen}{ + 0.0 } 	 & \textcolor{ForestGreen}{ + 1.9 } 	 & \textcolor{ForestGreen}{ + 0.6 } 	 & \textcolor{ForestGreen}{ + 2.1 } 	 & 93.1 	 & \textcolor{red}{  -0.4 } 	 & \textcolor{ForestGreen}{ + 4.7 } 	 & \textcolor{red}{  -0.7 }	 & \textcolor{ForestGreen}{ + 0.5 }\\
		50 & feeling warm                             	 & 96.5 	 & \textcolor{red}{  -1.0 } 	 & \textcolor{red}{  -0.3 } 	 & \textcolor{red}{  -2.2 } 	 & \textcolor{ForestGreen}{ + 1.8 } 	 & 90.5 	 & \textcolor{red}{  -2.2 } 	 & \textcolor{red}{  -0.7 } 	 & \textcolor{red}{  -0.7 }	 & \textcolor{ForestGreen}{ + 2.8 }\\
		51 & punching/slapping other person           	 & 96.5 	 & \textcolor{ForestGreen}{ + 0.3 } 	 & \textcolor{ForestGreen}{ + 0.6 } 	 & \textcolor{ForestGreen}{ + 1.6 } 	 & \textcolor{ForestGreen}{ + 2.2 } 	 & 92.8 	 & \textcolor{red}{  -2.9 } 	 & \textcolor{ForestGreen}{ + 2.5 } 	 & \textcolor{red}{  -2.9 }	 & \textcolor{ForestGreen}{ + 4.7 }\\
		52 & kicking other person                     	 & 98.7 	 & \textcolor{ForestGreen}{ + 0.6 } 	 & \textcolor{ForestGreen}{ + 1.0 } 	 & \textcolor{ForestGreen}{ + 0.3 } 	 & \textcolor{ForestGreen}{ + 1.3 } 	 & 97.5 	 & \textcolor{ForestGreen}{ + 0.7 } 	 & \textcolor{ForestGreen}{ + 1.1 } 	 & \textcolor{red}{  -0.7 }	 & \textcolor{ForestGreen}{ + 0.9 }\\
		53 & pushing other person                     	 & 96.8 	 & \textcolor{red}{  -2.2 } 	 & \textcolor{ForestGreen}{ + 1.9 } 	 & \textcolor{red}{  -5.1 } 	 & \textcolor{ForestGreen}{ + 0.8 } 	 & 95.3 	 & \textcolor{red}{  -0.7 } 	 & \textcolor{red}{  -0.7 } 	 & \textcolor{red}{  -1.1 }	 & \textcolor{ForestGreen}{ + 0.9 }\\
		54 & pat on back of other person              	 & 98.4 	 & \textcolor{red}{  -1.3 } 	 & \textcolor{red}{  -0.6 } 	 & \textcolor{red}{  -1.6 } 	 & \textcolor{red}{  -0.5 } 	 & 90.6 	 & \textcolor{red}{  -2.9 } 	 & \textcolor{ForestGreen}{ + 3.6 } 	 & \textcolor{ForestGreen}{ + 0.4 }	 & \textcolor{ForestGreen}{ + 5.9 }\\
		55 & point finger at the other person         	 & 99.0 	 & \textcolor{red}{  -0.3 } 	 & \textcolor{ForestGreen}{ + 0.3 } 	 & \textcolor{ForestGreen}{ + 0.3 }  & \textcolor{ForestGreen}{ + 0.8 } 	 & 98.5 	 & \textcolor{ForestGreen}{ + 0.4 } 	 & \textcolor{ForestGreen}{ + 0.7 } 	 & \textcolor{ForestGreen}{ + 0.7 }	 & \textcolor{ForestGreen}{ + 0.7 }\\
		56 & hugging other person                     	 & 96.5 	 & \textcolor{red}{  -0.6 } 	 & \textcolor{ForestGreen}{ + 2.2 } 	 & \textcolor{ForestGreen}{ + 1.3 }  & \textcolor{ForestGreen}{ + 2.5 } 	 & 90.9 	 & \textcolor{ForestGreen}{ + 1.8 } 	 & \textcolor{ForestGreen}{ + 4.0 } 	 & \textcolor{ForestGreen}{ + 2.2 }	 & \textcolor{ForestGreen}{ + 6.0 }\\
		57 & giving something to other person         	 & 96.2 	 & \textcolor{red}{  -0.9 } 	 & \textcolor{ForestGreen}{ + 1.9 } 	 & \textcolor{red}{  -2.5 } 	 & \textcolor{ForestGreen}{ + 1.9 } 	 & 96.4 	 & \textcolor{red}{  -1.8 } 	 & \textcolor{ForestGreen}{ + 1.5 } 	 & \textcolor{ForestGreen}{ + 0.0 }	 & \textcolor{red}{  -0.5 }\\
		58 & touch other persons pocket               	 & 98.1 	 & \textcolor{ForestGreen}{ + 0.6 } 	 & \textcolor{ForestGreen}{ + 1.3 } 	 & \textcolor{ForestGreen}{ + 0.0 } 	 & \textcolor{ForestGreen}{ + 1.3 } 	 & 95.3 	 & \textcolor{ForestGreen}{ + 2.9 } 	 & \textcolor{ForestGreen}{ + 1.4 } 	 & \textcolor{red}{  -1.1 }	 & \textcolor{ForestGreen}{ + 2.9 }\\
		59 & handshaking                              	 & 100.0 	 & \textcolor{red}{  -1.9 } 	 & \textcolor{red}{  -0.6 } 	 & \textcolor{red}{  -0.3 } 	 & \textcolor{red}{  -0.2 } 	 & 100.0 	 & \textcolor{ForestGreen}{ + 0.0 } 	 & \textcolor{ForestGreen}{ + 0.0 } 	 & \textcolor{ForestGreen}{ + 0.0 }	 & \textcolor{red}{  -0.2 }\\
		60 & walking towards each other               	 & 97.1 	 & \textcolor{red}{  -0.3 } 	 & \textcolor{ForestGreen}{ + 2.2 } 	 & \textcolor{ForestGreen}{ + 2.2 }  & \textcolor{ForestGreen}{ + 2.6 } 	 & 97.5 	 & \textcolor{ForestGreen}{ + 0.0 } 	 & \textcolor{ForestGreen}{ + 1.8 } 	 & \textcolor{ForestGreen}{ + 2.2 }	 & \textcolor{ForestGreen}{ + 2.4 }\\
		\hline
		&\textbf{ Average                                  }	 &\textbf{ 93.5 } 	 &\textbf{ \textcolor{ForestGreen}{ + 0.3 }} 	 &\textbf{ \textcolor{ForestGreen}{ + 4.0 }}  &\textbf{ \textcolor{ForestGreen}{ + 0.2 }} 	 &\textbf{ \textcolor{ForestGreen}{ + 4.0 }} 	 &\textbf{89.4 }	 &\textbf{ \textcolor{ForestGreen}{ + 0.4 }} 	 &\textbf{ \textcolor{ForestGreen}{ + 4.0 }} 	 &\textbf{ \textcolor{ForestGreen}{ + 0.9 }} 	 &\textbf{ \textcolor{ForestGreen}{ + 4.9 }}\\
		\bottomrule
	\end{tabular}
	\vspace{-10pt}
\end{table*}
Table \ref{table1} shows the class-wise performance comparison between our DeepActsNet (Conv+Graph) using body data only (baselines), using body+hands data (Model A), using body+flow data (Model B), using body+face+hands data (Model C), and using body+face+hands+bones+flow data (Model D) on the NTU60 dataset \cite{shahroudy2016ntu}. The results show that the addition of information from hands and optical flow improve accuracy for several action classes involving interactions with hands and face.
For instance, action classes involving interaction with hands such as ``pickup" ``clapping", ``making a phone call", ``writing", ``typing on a keyboard", and ``playing with phone" show considerable improvement in recognition accuracy for the models using additional hands data (Model A) compared to the baselines. The body joints data alone used in the baselines contain information of only two finger joints (``tip of the hand" and ``thumb"), making it difficult to capture subtle movements of the hands involved in these action classes. On the contrast, our Model A better encodes these actions by using more detailed information of finger joints, resulting in notable improvements in the accuracy as shown in Table \ref{table1}.
Furthermore, action classes involving interactions with the face such as ``eat meal", ``wear on glasses", ``shake head", ``headache", ``nausea", and ``touch head" show notable improvement in accuracy for the models using additional face data (Model A) compared to the baselines. 
Overall, for actions that have strong spatial-temporal joint correlations involving face and hands, our Deep Actions Stamps improve the accuracy of the models to greater extents (Model C and Model D). 
Table \ref{table1} also shows that action classes which involve less motion over time in face and hands compared to other actions and actions involving interactions between two people show deterioration in the recognition accuracy for models using additional face and hands data compared to the baselines. For instance, classes such as ``sneeze", ``punching other person", ``pushing other person", or ``kicking other person" present challenging cases where the models using additional face and hands data underperform compared to the baselines. This underperformance is mitigated by Model D which combines joints spatial data with explicit motion information in terms of optical flow, resulting in improved accuracy for most of the action classes. We experimentally found that the optical flow information estimated in terms of raw pixel spatial movements and orientations between adjacent frames is not affected by joints estimation errors due to occlusions and therefore provide better encoding of actions classes involving interactions between two people or occlusions. Overall, we see that our model using spatial and motion information from face, hands, and body (Model D) produces the best performance with improvements of around 4.6\% and 9.1\% in the Cross-View (CV) and Cross-Subject (CS) settings, respectively compared to the baselines.
\begin{figure*}[htbp]
	\begin{center}
		\includegraphics[trim=0.5cm 0.2cm 5.0cm 5cm,clip,width=0.6\linewidth,keepaspectratio]{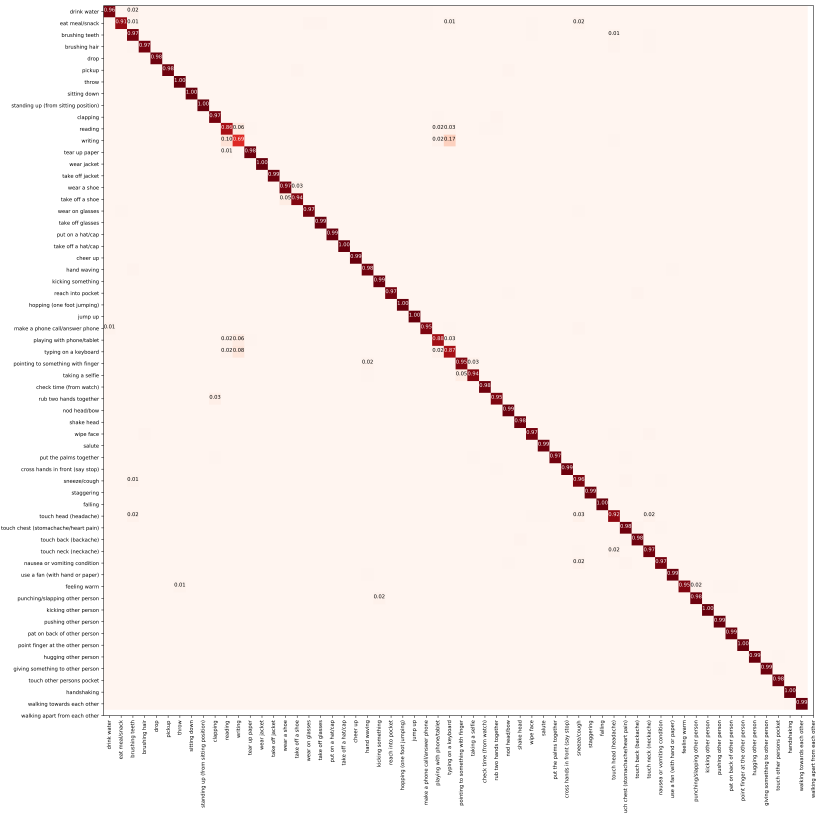}
		\vspace{-5pt}
		\caption{Confusion matrix produced by our DeepActsNet (Conv+Graph) on the Cross Subject setting of the NTU60 dataset \cite{shahroudy2016ntu} , using the joint data only. Similar action pairs such as “writing” vs.  “typing on a keyboard”, and “reading” vs. ”playing with phone/tablet” are more frequently misclassified by our model trained on body joints information alone. Figure best viewed with color and zoomed in.}
		\label{fig_cm1}
	\end{center}
	\vspace{-20pt}
\end{figure*}
\begin{figure*}[htbp]
	\begin{center}
		\includegraphics[trim=0.5cm 0.2cm 5.0cm 5cm,clip,width=0.6\linewidth,keepaspectratio]{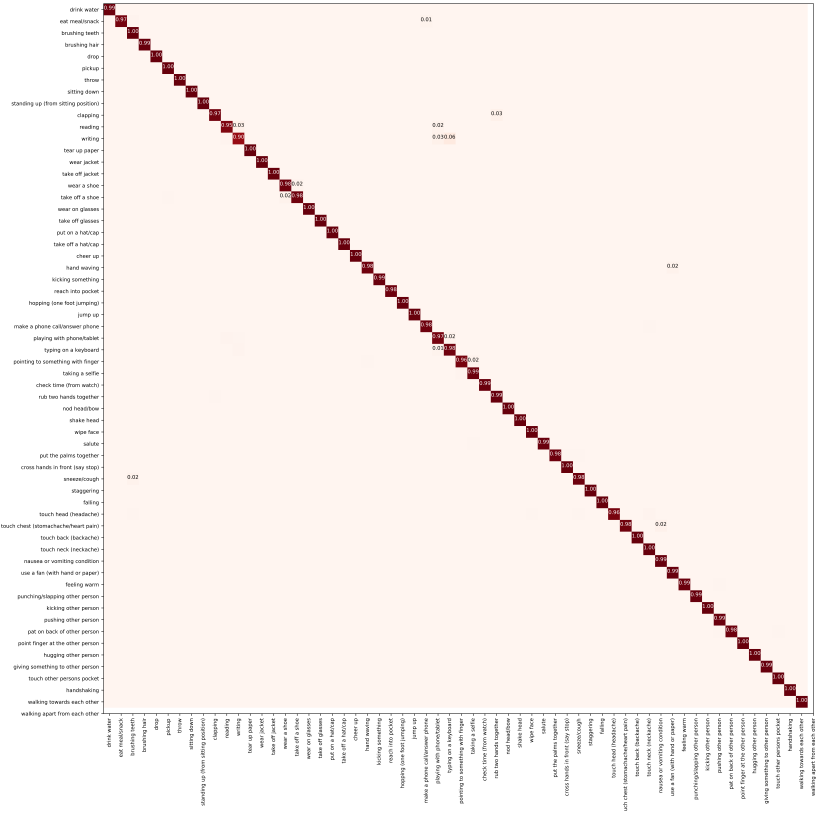}
		\vspace{-5pt}
		\caption{Confusion matrix produced by our DeepActsNet (Conv+Graph) on the Cross Subject setting of the NTU60 dataset \cite{shahroudy2016ntu}, using body+face+hands data. The results show that the model trained using the additional information from face and hands better discriminate actions involving similar hand movements and produce less confusions compared to the models trained using body joints information alone (Fig. \ref{fig_cm1}). Figure best viewed with color and zoomed in.}
		\label{fig_cm2}
	\end{center}
	\vspace{-20pt}
\end{figure*}
\begin{figure*}[htbp]
	\begin{center}
		\includegraphics[trim=0.5cm 0.2cm 5.0cm 5cm,clip,width=0.6\linewidth,keepaspectratio]{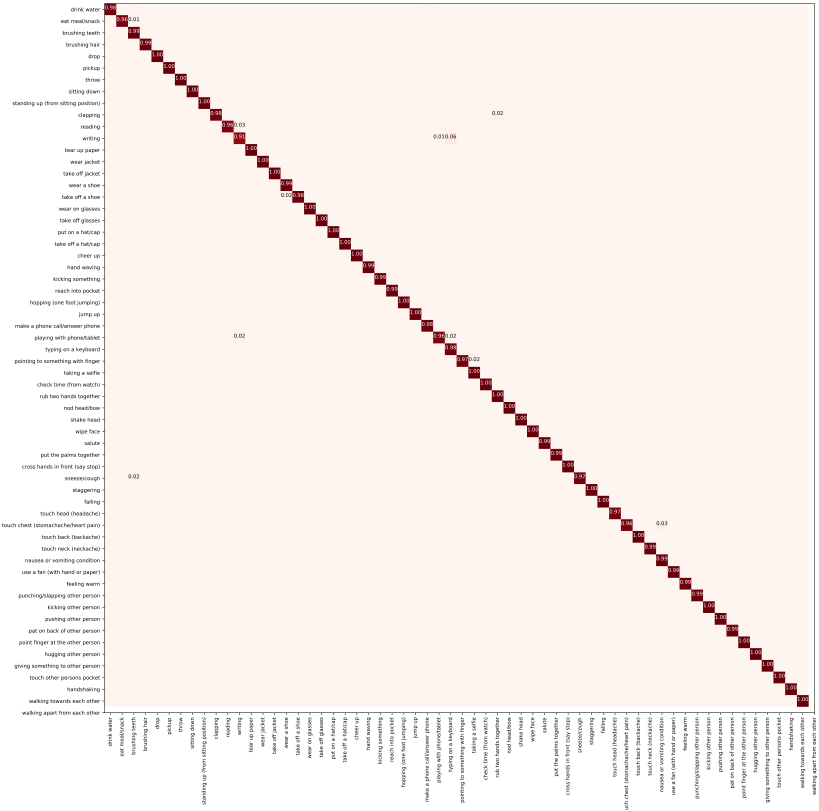}
		\vspace{-5pt}
		\caption{Confusion matrix produced by our DeepActsNet (Conv+Graph) on the Cross Subject setting of the NTU60 dataset \cite{shahroudy2016ntu}, using body+face+hands+flow data. The results show that the model trained using both spatial and motion information from face and hands produces the least confusions for most of the action classes compared to the models trained using body joints information alone (Fig. \ref{fig_cm1}). Figure best viewed with color and zoomed in.}
		\label{fig_cm3}
	\end{center}
	\vspace{-20pt}
\end{figure*}
\subsubsection{Occlusions and false positives}
Fig. \ref{fig_errors} shows examples where skeleton data contain missing joints due to occlusions. For these cases, methods solely working on body-only, hands-only, or face-only joints would fail drastically. Therefore we built Deep Action Stamps to harnesses holistic information from all visible joints in addition to optical flow information resulting in improved recognition accuracy compared to the individual modalities (see Table \ref{table_ablation}). To minimize errors in pose estimation, we trained the pose models with augmented data containing occluded body parts which improved their robustness to occlusions in real-world scenes. Furthermore, we employed a method to filter and consider only high quality joints based on joints confidence value to minimize the effect of false positives in pose data on action recognition. It is also to be noted that pose estimation was not the main focus of the paper and therefore can be expanded with domain-specific methods for more accurate pose estimation over other datasets.
\subsubsection{Ablation Study of DeepActsNet}
Here we examine the significance of the fusion of convolutional and structural features in DeepActsNet for activity recognition. 
Table \ref{table_ablation}(M-Q) show that the model using Conv+Graph streams consistently improved the accuracy for all the datasets compared to the Conv-only or Graph-only streams. For instance, DeepActsNet with Conv+Graph streams using body joints only (row-M in Table \ref{table_ablation}) yielded improvements of 3.6\% and 2.3\% in the cross-subject and cross-view accuracy on the NTU60 dataset, respectively compared to the model with only Conv stream (row-D in Table \ref{table_ablation}). Compared to the model with only Graph-stream, DeepActsNet with Conv+Graph streams produced improvements of 4.2\% and 4.8\% in the cross-subject and cross-view accuracy, respectively (row-J in Table \ref{table_ablation}). 
\\
\indent
The benefits of the fusion of convolutional and structural features can also be seen for the cases of models that were trained using multi-modal data. For instance, DeepActsNet with Conv+Graph streams using multi-modal data (row-Q in Table \ref{table_ablation}) yielded improvements of 1.6\% and 0.4\% in the cross-subject and cross-view accuracy on the NTU60 dataset, respectively compared to the models with only Conv-stream (row-F in Table \ref{table_ablation}). Furthermore, DeepActsNet with Conv+Graph streams produced improvements of 1.4\% and 1.8\% in the cross-subject and cross-view accuracy, respectively compared to the models with only Graph-stream (row-L in Table \ref{table_ablation}).
Fig. \ref{fig_cm1}, Fig. \ref{fig_cm2}, and Fig. \ref{fig_cm3} show confusion matrices produced by DeepActsNet (Conv+Graph) using body joints information alone, using body+face+hands data, and using body+face+hands+flow data, respectively on the NTU60 \cite{shahroudy2016ntu}. The comparison shows that the addition of information from face, hands, and optical flow results in reduction of confusions for most of the confusing classes that were miss-classified by the model using body joint data alone (Fig. \ref{fig_cm1}). For instance, action classes such as ``writing" and ``reading" are confused with classes such as ``typing on a keyboard" and ``playing with phone" for the model using body joints information alone. This is because, the distinguishing features for these calsses include minor hand and fore-arm movements but they are ineffectively captured by limited joint information in the body skeleton data (which represent hands with only ``tip of the hand" and ``thumb"). On the contrary, our Deep Action Stamps encode more detailed information from hands in terms of 21 finger joints information for each hand, making it easy to capture subtle movements of the hands involved in these challenging action classes. Other challenges that are evident in the body skeleton data include missing joints information or noisy joints estimation due to occlusions. This results in different actions having very similar poses as shown by the confusion matrix in Fig. \ref{fig_cm1}. For these challenging cases, the combination spatial and motion ifnormation from face and hands provides a more effective encoding of actions compared to the body skeleton data alone, thereby enabling the models trained using Deep Actions Stamps produce considerably less confusions as shown in Fig. \ref{fig_cm2} and Fig. \ref{fig_cm3}.
\begin{table*}[htbp]
	\caption{Details of the architectures of different variants of the proposed Enhanced Convolutional Graph Network (ECGN). The input tensor for the Conv branch denote feature channels $C$, and width $W$ and height $H$ of the feature channels. The input tensor for the Graph branch are represented by feature channels $C$, temporal length $T$, number of joints $N_{j}$, and number of people $N_{p}$. In this work, we used $T=60$ and $N_{p}=2$ for NTU60 \cite{shahroudy2016ntu} , NTU120 \cite{liu2019ntu}, and SYSU \cite{hu2015jointly} datasets. The width $W$ and height $W$ parameters for the feature channels were set to $224$. The number of joints $N_{j}$ were set to $25$ for body keypoints, $68$ for facial keypoints, $42$ for the keypoints of two hands, and $25$ for bones and optical flow information. For all experiments, we used $C=3$ feature channels. The output tensors are $M\times 1-$ dimensional where $M$ denotes the number of target classes.}
	\vspace{-10pt}
	\label{table_network}
	\centering
	\setlength\tabcolsep{3pt}
	\begin{tabular}{@{}l|cc|cc|cc@{}}
		\toprule
		&\multicolumn{2}{c|}{ECGN - Deep version}&\multicolumn{2}{c|}{ECGN - Lite version}& \multicolumn{2}{c}{ECGN - Tiny version}\\
		& \multicolumn{2}{c|}{Parameters: 67.0 M, GFLOPs: 3.3} & \multicolumn{2}{c|}{Parameters: 64.7 M, GFLOPs: 0.7}&  \multicolumn{2}{c}{Parameters: 4.9 M, GFLOPs: 0.6}\\ 
		{Layer name} &Output Size&No. of layers&Output Size&No. of layers&Output Size&No. of layers\\					
		\midrule
		Input& $C\times W\times H$&-& $C\times W\times H$&-& $C\times W\times H$&-\\
		Conv $3\times3$ &$32\times 112\times 112$&4&$32\times 112\times 112$&4&$32\times 112\times 112$&2\\
		MBConv  $3\times3 $&$48\times 56\times 56$&7&$48\times 56\times 56$&7&$24\times 56\times 56$&2\\		
		MBConv $5\times5 $&$80\times 28\times 28$&7&$80\times 28\times 28$&7&$40\times 28\times 28$&2\\				
		MBConv $5\times5$ &$160\times 14\times 14$&10&$160\times 14\times 14$&10&$80\times 14\times 14$&3\\
		MBConv $5\times5$ &$224\times 14\times 14$&10&$224\times 14\times 14$&10&$80\times 14\times 14$&3\\		
		MBConv $5\times5 $&$384\times 7\times 7$&13&$384\times 7\times 7$&13&$192\times 7\times 7$&1\\		
		MBConv $3\times3$ &$640\times 7\times 7$&4 &$640\times 7\times 7$&4&$320\times 7\times 7$&1\\
		Conv$ 1\times1 $&$2560\times 7\times 7$&1&$2560\times 7\times 7$&1&$1280\times 7\times 7$&1\\
		Average Pool  &$2560\times 1$&1&$2560\times 1$&1&$1280\times 1$&1\\
		$fc_{conv}$ &$M\times 1$&1&$M\times 1$&1&$M\times 1$&1\\
		\hline
		Input&$C\times T\times N_{j}\times N_{p}$&-&$C\times T\times N_{j}\times N_{p}$&-&$C\times T\times N_{j}\times N_{p}$&-\\
		GCN0&$64\times 60\times 25$&1&$32\times 60\times 25$&1&$32\times 60\times 25$&1\\
		GCN1&$64\times 60\times 25$&1&$32\times 60\times 25$&1&$32\times 60\times 25$&1\\
		GCN2&$64\times 60\times 25$&1&$32\times 60\times 25$&1&$32\times 60\times 25$&1\\
		GCN3&$64\times 60\times 25$&1&$32\times 60\times 25$&1&$32\times 60\times 25$&1\\
		GCN4&$128\times 30\times 25$&1&$64\times 30\times 25$&1&$64\times 30\times 25$&1\\
		GCN5&$128\times 30\times 25$&1&$64\times 30\times 25$&1&$64\times 30\times 25$&1\\
		GCN6&$128\times 30\times 25$&1&$64\times 30\times 25$&1&$64\times 30\times 25$&1\\
		GCN7&$256\times 15\times 25$&1&$128\times 15\times 25$&1&$128\times 15\times 25$&1\\
		GCN8&$256\times 15\times 25$&1&$128\times 15\times 25$&1&$128\times 15\times 25$&1\\
		GCN9&$256\times 15\times 25$&1&$128\times 15\times 25$&1&$128\times 15\times 25$&1\\
		Average Pool  &$256\times 15\times 1$&1&$128\times 15\times 1$&1&$128\times 15\times 1$&1\\
		Average Pool  &$256\times 1$&1&$128\times 1$&1&$128\times 1$&1\\
		$fc_{graph}$ &$M\times 1$&1&$M\times 1$&1&$M\times 1$&1\\
		\bottomrule
	\end{tabular}
	\vspace{0pt}
\end{table*}
\begin{table*}[htbp]
	\caption{Improvements in class-wise accuracy of our models for the 60 classes of the NTU60 dataset \cite{liu2019ntu} on the Cross-Subject (CS) setting, using Conv-stream only (Model A), using Graph-stream only (Model B), using Conv+Graph streams (Model C), and using Conv+Graph streams with shallow layers (Model D). We used the DGNN model of \cite{shi2019skeleton} as the performance baseline. Values in green and red are accuracy differences compared to the baselines. The classes are sorted in the ascending order with respect to the baseline accuracy.}
	\vspace{-10pt}
	\label{table_ntu}
	\centering
	\setlength\tabcolsep{5.0pt}
	\begin{tabular}{@{}clcccccccccc@{}}
		\toprule   
		&&Baseline (DGNN \cite{shi2019skeleton})&	{Model A} &Model B&Model C&Model D\\			
		&&GFLOPs: 126.8&GFLOPs: 0.4&GFLOPs: 16.3&GFLOPs: 16.7&GFLOPs: 7.1\\
		&{Action Class}&CS (\%) &CS (\%)&{CS (\%)} &CS (\%)&CS (\%)\\					
		\midrule
		1 & writing                                  	 & 54.4 	 & 73.9 ( \textcolor{ForestGreen}{ + 19.5 } ) 	 & 58.8 ( \textcolor{ForestGreen}{ + 4.4 } ) 	 & 70.7 ( \textcolor{ForestGreen}{ + 16.3 } ) 	 & 74.3 ( \textcolor{ForestGreen}{ + 19.9 } ) \\
		2 & typing on a keyboard                     	 & 67.3 	 & 83.3 ( \textcolor{ForestGreen}{ + 16.0 } ) 	 & 86.2 ( \textcolor{ForestGreen}{ + 18.9 } ) 	 & 83.0 ( \textcolor{ForestGreen}{ + 15.7 } ) 	 & 86.9 ( \textcolor{ForestGreen}{ + 19.6 } ) \\
		3 & eat meal/snack                           	 & 67.6 	 & 75.3 ( \textcolor{ForestGreen}{ + 7.7 } ) 	 & 83.3 ( \textcolor{ForestGreen}{ + 15.7 } ) 	 & 87.3 ( \textcolor{ForestGreen}{ + 19.7 } ) 	 & 80.7 ( \textcolor{ForestGreen}{ + 13.1 } ) \\
		4 & reading                                  	 & 68.5 	 & 59.7 ( \textcolor{red}{  -8.8 } ) 	 & 68.9 ( \textcolor{ForestGreen}{ + 0.4 } ) 	 & 76.4 ( \textcolor{ForestGreen}{ + 7.9 } ) 	 & 72.2 ( \textcolor{ForestGreen}{ + 3.7 } ) \\
		5 & playing with phone/tablet                	 & 72.4 	 & 62.9 ( \textcolor{red}{  -9.5 } ) 	 & 82.5 ( \textcolor{ForestGreen}{ + 10.1 } ) 	 & 78.3 ( \textcolor{ForestGreen}{ + 5.9 } ) 	 & 78.9 ( \textcolor{ForestGreen}{ + 6.5 } ) \\
		6 & sneeze/cough                             	 & 75.7 	 & 72.8 ( \textcolor{red}{  -2.9 } ) 	 & 82.6 ( \textcolor{ForestGreen}{ + 6.9 } ) 	 & 87.8 ( \textcolor{ForestGreen}{ + 12.1 } ) 	 & 84.4 ( \textcolor{ForestGreen}{ + 8.7 } ) \\
		7 & pointing to something with finger        	 & 81.2 	 & 80.8 ( \textcolor{red}{  -0.4 } ) 	 & 87.3 ( \textcolor{ForestGreen}{ + 6.1 } ) 	 & 91.3 ( \textcolor{ForestGreen}{ + 10.1 } ) 	 & 86.6 ( \textcolor{ForestGreen}{ + 5.4 } ) \\
		8 & clapping                                 	 & 82.8 	 & 78.4 ( \textcolor{red}{  -4.4 } ) 	 & 92.7 ( \textcolor{ForestGreen}{ + 9.9 } ) 	 & 92.4 ( \textcolor{ForestGreen}{ + 9.6 } ) 	 & 86.8 ( \textcolor{ForestGreen}{ + 4.0 } ) \\
		9 & touch head (headache)                    	 & 83.3 	 & 87.7 ( \textcolor{ForestGreen}{ + 4.4 } ) 	 & 86.6 ( \textcolor{ForestGreen}{ + 3.3 } ) 	 & 84.4 ( \textcolor{ForestGreen}{ + 1.1 } ) 	 & 87.7 ( \textcolor{ForestGreen}{ + 4.4 } ) \\
		10 & reach into pocket                        	 & 83.9 	 & 89.1 ( \textcolor{ForestGreen}{ + 5.2 } ) 	 & 88.0 ( \textcolor{ForestGreen}{ + 4.1 } ) 	 & 91.9 ( \textcolor{ForestGreen}{ + 8.0 } ) 	 & 88.3 ( \textcolor{ForestGreen}{ + 4.4 } ) \\
		11 & wear a shoe                              	 & 83.9 	 & 91.2 ( \textcolor{ForestGreen}{ + 7.3 } ) 	 & 87.9 ( \textcolor{ForestGreen}{ + 4.0 } ) 	 & 91.5 ( \textcolor{ForestGreen}{ + 7.6 } ) 	 & 81.7 ( \textcolor{red}{  -2.2 } ) \\
		12 & nausea or vomiting condition             	 & 84.0 	 & 94.5 ( \textcolor{ForestGreen}{ + 10.5 } ) 	 & 92.0 ( \textcolor{ForestGreen}{ + 8.0 } ) 	 & 91.7 ( \textcolor{ForestGreen}{ + 7.7 } ) 	 & 91.3 ( \textcolor{ForestGreen}{ + 7.3 } ) \\
		13 & drink water                              	 & 85.8 	 & 95.6 ( \textcolor{ForestGreen}{ + 9.8 } ) 	 & 89.8 ( \textcolor{ForestGreen}{ + 4.0 } ) 	 & 94.4 ( \textcolor{ForestGreen}{ + 8.6 } ) 	 & 95.6 ( \textcolor{ForestGreen}{ + 9.8 } ) \\
		14 & touch neck (neckache)                    	 & 86.2 	 & 89.9 ( \textcolor{ForestGreen}{ + 3.7 } ) 	 & 90.9 ( \textcolor{ForestGreen}{ + 4.7 } ) 	 & 94.4 ( \textcolor{ForestGreen}{ + 8.2 } ) 	 & 92.0 ( \textcolor{ForestGreen}{ + 5.8 } ) \\
		15 & make a phone call/answer phone           	 & 86.2 	 & 90.9 ( \textcolor{ForestGreen}{ + 4.7 } ) 	 & 94.5 ( \textcolor{ForestGreen}{ + 8.3 } ) 	 & 93.7 ( \textcolor{ForestGreen}{ + 7.5 } ) 	 & 96.0 ( \textcolor{ForestGreen}{ + 9.8 } ) \\
		16 & rub two hands together                   	 & 86.6 	 & 93.5 ( \textcolor{ForestGreen}{ + 6.9 } ) 	 & 88.0 ( \textcolor{ForestGreen}{ + 1.4 } ) 	 & 91.9 ( \textcolor{ForestGreen}{ + 5.3 } ) 	 & 96.7 ( \textcolor{ForestGreen}{ + 10.1 } ) \\
		17 & take off a shoe                          	 & 87.2 	 & 88.0 ( \textcolor{ForestGreen}{ + 0.8 } ) 	 & 68.6 ( \textcolor{red}{  -18.6 } ) 	 & 89.3 ( \textcolor{ForestGreen}{ + 2.1 } ) 	 & 90.9 ( \textcolor{ForestGreen}{ + 3.7 } ) \\
		18 & wipe face                                	 & 87.3 	 & 91.3 ( \textcolor{ForestGreen}{ + 4.0 } ) 	 & 97.1 ( \textcolor{ForestGreen}{ + 9.8 } ) 	 & 93.8 ( \textcolor{ForestGreen}{ + 6.5 } ) 	 & 94.2 ( \textcolor{ForestGreen}{ + 6.9 } ) \\
		19 & brushing teeth                           	 & 87.5 	 & 91.9 ( \textcolor{ForestGreen}{ + 4.4 } ) 	 & 91.6 ( \textcolor{ForestGreen}{ + 4.1 } ) 	 & 92.3 ( \textcolor{ForestGreen}{ + 4.8 } ) 	 & 93.0 ( \textcolor{ForestGreen}{ + 5.5 } ) \\
		20 & pat on back of other person              	 & 87.7 	 & 95.3 ( \textcolor{ForestGreen}{ + 7.6 } ) 	 & 92.0 ( \textcolor{ForestGreen}{ + 4.3 } ) 	 & 96.5 ( \textcolor{ForestGreen}{ + 8.8 } ) 	 & 92.4 ( \textcolor{ForestGreen}{ + 4.7 } ) \\
		21 & check time (from watch)                  	 & 88.4 	 & 95.3 ( \textcolor{ForestGreen}{ + 6.9 } ) 	 & 92.4 ( \textcolor{ForestGreen}{ + 4.0 } ) 	 & 96.6 ( \textcolor{ForestGreen}{ + 8.2 } ) 	 & 94.9 ( \textcolor{ForestGreen}{ + 6.5 } ) \\
		22 & taking a selfie                          	 & 89.1 	 & 91.7 ( \textcolor{ForestGreen}{ + 2.6 } ) 	 & 96.0 ( \textcolor{ForestGreen}{ + 6.9 } ) 	 & 92.2 ( \textcolor{ForestGreen}{ + 3.1 } ) 	 & 95.3 ( \textcolor{ForestGreen}{ + 6.2 } ) \\
		23 & feeling warm                             	 & 89.8 	 & 92.7 ( \textcolor{ForestGreen}{ + 2.9 } ) 	 & 90.1 ( \textcolor{ForestGreen}{ + 0.3 } ) 	 & 93.4 ( \textcolor{ForestGreen}{ + 3.6 } ) 	 & 89.8 ( \textcolor{red}{  -0.0 } ) \\
		24 & drop                                     	 & 89.8 	 & 96.4 ( \textcolor{ForestGreen}{ + 6.6 } ) 	 & 89.5 ( \textcolor{red}{  -0.3 } ) 	 & 97.4 ( \textcolor{ForestGreen}{ + 7.6 } ) 	 & 95.6 ( \textcolor{ForestGreen}{ + 5.8 } ) \\
		25 & salute                                   	 & 89.9 	 & 96.4 ( \textcolor{ForestGreen}{ + 6.5 } ) 	 & 94.9 ( \textcolor{ForestGreen}{ + 5.0 } ) 	 & 96.0 ( \textcolor{ForestGreen}{ + 6.1 } ) 	 & 96.4 ( \textcolor{ForestGreen}{ + 6.5 } ) \\
		26 & touch chest (stomachache/heart pain)     	 & 90.2 	 & 86.2 ( \textcolor{red}{  -4.0 } ) 	 & 91.7 ( \textcolor{ForestGreen}{ + 1.5 } ) 	 & 94.0 ( \textcolor{ForestGreen}{ + 3.8 } ) 	 & 93.5 ( \textcolor{ForestGreen}{ + 3.3 } ) \\
		27 & tear up paper                            	 & 90.4 	 & 96.7 ( \textcolor{ForestGreen}{ + 6.3 } ) 	 & 95.2 ( \textcolor{ForestGreen}{ + 4.8 } ) 	 & 96.6 ( \textcolor{ForestGreen}{ + 6.2 } ) 	 & 96.7 ( \textcolor{ForestGreen}{ + 6.3 } ) \\
		28 & brushing hair                            	 & 90.8 	 & 90.8 ( \textcolor{ForestGreen}{ + 0.0 } ) 	 & 90.1 ( \textcolor{red}{  -0.7 } ) 	 & 94.5 ( \textcolor{ForestGreen}{ + 3.7 } ) 	 & 93.8 ( \textcolor{ForestGreen}{ + 3.0 } ) \\
		29 & wear on glasses                          	 & 91.6 	 & 93.0 ( \textcolor{ForestGreen}{ + 1.4 } ) 	 & 93.8 ( \textcolor{ForestGreen}{ + 2.2 } ) 	 & 95.8 ( \textcolor{ForestGreen}{ + 4.2 } ) 	 & 96.7 ( \textcolor{ForestGreen}{ + 5.1 } ) \\
		30 & use a fan (with hand or paper)           	 & 91.6 	 & 93.1 ( \textcolor{ForestGreen}{ + 1.5 } ) 	 & 95.6 ( \textcolor{ForestGreen}{ + 4.0 } ) 	 & 93.6 ( \textcolor{ForestGreen}{ + 2.0 } ) 	 & 96.0 ( \textcolor{ForestGreen}{ + 4.4 } ) \\
		31 & pickup                                   	 & 91.6 	 & 96.0 ( \textcolor{ForestGreen}{ + 4.4 } ) 	 & 98.9 ( \textcolor{ForestGreen}{ + 7.3 } ) 	 & 97.8 ( \textcolor{ForestGreen}{ + 6.2 } ) 	 & 97.1 ( \textcolor{ForestGreen}{ + 5.5 } ) \\
		32 & take off glasses                         	 & 91.6 	 & 96.7 ( \textcolor{ForestGreen}{ + 5.1 } ) 	 & 97.1 ( \textcolor{ForestGreen}{ + 5.5 } ) 	 & 95.7 ( \textcolor{ForestGreen}{ + 4.1 } ) 	 & 97.1 ( \textcolor{ForestGreen}{ + 5.5 } ) \\
		33 & punching/slapping other person           	 & 92.0 	 & 94.9 ( \textcolor{ForestGreen}{ + 2.9 } ) 	 & 92.8 ( \textcolor{ForestGreen}{ + 0.8 } ) 	 & 97.4 ( \textcolor{ForestGreen}{ + 5.4 } ) 	 & 94.9 ( \textcolor{ForestGreen}{ + 2.9 } ) \\
		34 & giving something to other person         	 & 92.0 	 & 96.7 ( \textcolor{ForestGreen}{ + 4.7 } ) 	 & 96.7 ( \textcolor{ForestGreen}{ + 4.7 } ) 	 & 95.9 ( \textcolor{ForestGreen}{ + 3.9 } ) 	 & 97.8 ( \textcolor{ForestGreen}{ + 5.8 } ) \\
		35 & hand waving                              	 & 92.3 	 & 94.2 ( \textcolor{ForestGreen}{ + 1.9 } ) 	 & 92.7 ( \textcolor{ForestGreen}{ + 0.4 } ) 	 & 94.7 ( \textcolor{ForestGreen}{ + 2.4 } ) 	 & 93.1 ( \textcolor{ForestGreen}{ + 0.8 } ) \\
		36 & pushing other person                     	 & 92.4 	 & 90.9 ( \textcolor{red}{  -1.5 } ) 	 & 97.1 ( \textcolor{ForestGreen}{ + 4.7 } ) 	 & 96.2 ( \textcolor{ForestGreen}{ + 3.8 } ) 	 & 97.8 ( \textcolor{ForestGreen}{ + 5.4 } ) \\
		37 & put the palms together                   	 & 92.4 	 & 92.8 ( \textcolor{ForestGreen}{ + 0.4 } ) 	 & 96.0 ( \textcolor{ForestGreen}{ + 3.6 } ) 	 & 96.6 ( \textcolor{ForestGreen}{ + 4.2 } ) 	 & 94.9 ( \textcolor{ForestGreen}{ + 2.5 } ) \\
		38 & hugging other person                     	 & 93.1 	 & 94.2 ( \textcolor{ForestGreen}{ + 1.1 } ) 	 & 96.0 ( \textcolor{ForestGreen}{ + 2.9 } ) 	 & 96.9 ( \textcolor{ForestGreen}{ + 3.8 } ) 	 & 94.6 ( \textcolor{ForestGreen}{ + 1.5 } ) \\
		39 & throw                                    	 & 94.2 	 & 94.9 ( \textcolor{ForestGreen}{ + 0.7 } ) 	 & 92.0 ( \textcolor{red}{  -2.2 } ) 	 & 97.8 ( \textcolor{ForestGreen}{ + 3.6 } ) 	 & 94.2 ( \textcolor{red}{  -0.0 } ) \\
		40 & sitting down                             	 & 94.5 	 & 98.5 ( \textcolor{ForestGreen}{ + 4.0 } ) 	 & 97.4 ( \textcolor{ForestGreen}{ + 2.9 } ) 	 & 99.1 ( \textcolor{ForestGreen}{ + 4.6 } ) 	 & 98.5 ( \textcolor{ForestGreen}{ + 4.0 } ) \\
		41 & shake head                               	 & 94.5 	 & 99.6 ( \textcolor{ForestGreen}{ + 5.1 } ) 	 & 99.3 ( \textcolor{ForestGreen}{ + 4.8 } ) 	 & 99.6 ( \textcolor{ForestGreen}{ + 5.1 } ) 	 & 100.0 ( \textcolor{ForestGreen}{ + 5.5 } ) \\
		42 & nod head/bow                             	 & 94.6 	 & 92.4 ( \textcolor{red}{  -2.2 } ) 	 & 98.9 ( \textcolor{ForestGreen}{ + 4.3 } ) 	 & 96.8 ( \textcolor{ForestGreen}{ + 2.2 } ) 	 & 99.3 ( \textcolor{ForestGreen}{ + 4.7 } ) \\
		43 & cross hands in front (say stop)          	 & 94.9 	 & 96.4 ( \textcolor{ForestGreen}{ + 1.5 } ) 	 & 95.3 ( \textcolor{ForestGreen}{ + 0.4 } ) 	 & 97.8 ( \textcolor{ForestGreen}{ + 2.9 } ) 	 & 97.8 ( \textcolor{ForestGreen}{ + 2.9 } ) \\
		44 & touch back (backache)                    	 & 94.9 	 & 96.7 ( \textcolor{ForestGreen}{ + 1.8 } ) 	 & 97.1 ( \textcolor{ForestGreen}{ + 2.2 } ) 	 & 96.1 ( \textcolor{ForestGreen}{ + 1.2 } ) 	 & 97.1 ( \textcolor{ForestGreen}{ + 2.2 } ) \\
		45 & take off a hat/cap                       	 & 94.9 	 & 99.6 ( \textcolor{ForestGreen}{ + 4.7 } ) 	 & 99.3 ( \textcolor{ForestGreen}{ + 4.4 } ) 	 & 99.5 ( \textcolor{ForestGreen}{ + 4.6 } ) 	 & 99.6 ( \textcolor{ForestGreen}{ + 4.7 } ) \\
		46 & put on a hat/cap                         	 & 95.2 	 & 97.4 ( \textcolor{ForestGreen}{ + 2.2 } ) 	 & 96.3 ( \textcolor{ForestGreen}{ + 1.1 } ) 	 & 98.9 ( \textcolor{ForestGreen}{ + 3.7 } ) 	 & 97.8 ( \textcolor{ForestGreen}{ + 2.6 } ) \\
		47 & cheer up                                 	 & 95.3 	 & 93.8 ( \textcolor{red}{  -1.5 } ) 	 & 97.4 ( \textcolor{ForestGreen}{ + 2.1 } ) 	 & 97.4 ( \textcolor{ForestGreen}{ + 2.1 } ) 	 & 98.5 ( \textcolor{ForestGreen}{ + 3.2 } ) \\
		48 & walking towards each other               	 & 96.0 	 & 99.3 ( \textcolor{ForestGreen}{ + 3.3 } ) 	 & 99.6 ( \textcolor{ForestGreen}{ + 3.6 } ) 	 & 99.8 ( \textcolor{ForestGreen}{ + 3.8 } ) 	 & 98.6 ( \textcolor{ForestGreen}{ + 2.6 } ) \\
		49 & kicking something                        	 & 96.4 	 & 96.0 ( \textcolor{red}{  -0.4 } ) 	 & 94.2 ( \textcolor{red}{  -2.2 } ) 	 & 96.1 ( \textcolor{red}{  -0.3 } ) 	 & 98.2 ( \textcolor{ForestGreen}{ + 1.8 } ) \\
		50 & take off jacket                          	 & 96.7 	 & 98.2 ( \textcolor{ForestGreen}{ + 1.5 } ) 	 & 95.7 ( \textcolor{red}{  -1.0 } ) 	 & 97.6 ( \textcolor{ForestGreen}{ + 0.9 } ) 	 & 99.3 ( \textcolor{ForestGreen}{ + 2.6 } ) \\
		51 & touch other persons pocket               	 & 97.1 	 & 96.7 ( \textcolor{red}{  -0.4 } ) 	 & 97.5 ( \textcolor{ForestGreen}{ + 0.4 } ) 	 & 98.2 ( \textcolor{ForestGreen}{ + 1.1 } ) 	 & 97.8 ( \textcolor{ForestGreen}{ + 0.7 } ) \\
		52 & kicking other person                     	 & 97.1 	 & 98.6 ( \textcolor{ForestGreen}{ + 1.5 } ) 	 & 97.1 ( \textcolor{ForestGreen}{ + 0.0 } ) 	 & 98.4 ( \textcolor{ForestGreen}{ + 1.3 } ) 	 & 99.6 ( \textcolor{ForestGreen}{ + 2.5 } ) \\
		53 & hopping (one foot jumping)               	 & 97.5 	 & 98.9 ( \textcolor{ForestGreen}{ + 1.4 } ) 	 & 98.5 ( \textcolor{ForestGreen}{ + 1.0 } ) 	 & 99.3 ( \textcolor{ForestGreen}{ + 1.8 } ) 	 & 98.9 ( \textcolor{ForestGreen}{ + 1.4 } ) \\
		54 & falling                                  	 & 97.8 	 & 99.6 ( \textcolor{ForestGreen}{ + 1.8 } ) 	 & 97.1 ( \textcolor{red}{  -0.7 } ) 	 & 99.1 ( \textcolor{ForestGreen}{ + 1.3 } ) 	 & 100.0 ( \textcolor{ForestGreen}{ + 2.2 } ) \\
		55 & handshaking                              	 & 97.8 	 & 100.0 ( \textcolor{ForestGreen}{ + 2.2 } ) 	 & 100.0 ( \textcolor{ForestGreen}{ + 2.2 } ) 	 & 99.8 ( \textcolor{ForestGreen}{ + 2.0 } ) 	 & 100.0 ( \textcolor{ForestGreen}{ + 2.2 } ) \\
		56 & standing up (from sitting position)      	 & 98.2 	 & 98.9 ( \textcolor{ForestGreen}{ + 0.7 } ) 	 & 99.3 ( \textcolor{ForestGreen}{ + 1.1 } ) 	 & 99.3 ( \textcolor{ForestGreen}{ + 1.1 } ) 	 & 98.9 ( \textcolor{ForestGreen}{ + 0.7 } ) \\
		57 & point finger at the other person         	 & 98.5 	 & 99.3 ( \textcolor{ForestGreen}{ + 0.8 } ) 	 & 98.5 ( \textcolor{ForestGreen}{ + 0.0 } ) 	 & 99.3 ( \textcolor{ForestGreen}{ + 0.8 } ) 	 & 98.9 ( \textcolor{ForestGreen}{ + 0.4 } ) \\
		58 & wear jacket                              	 & 98.9 	 & 98.9 ( \textcolor{ForestGreen}{ + 0.0 } ) 	 & 98.9 ( \textcolor{ForestGreen}{ + 0.0 } ) 	 & 98.0 ( \textcolor{red}{  -0.9 } ) 	 & 98.9 ( \textcolor{ForestGreen}{ + 0.0 } ) \\
		59 & jump up                                  	 & 99.3 	 & 100.0 ( \textcolor{ForestGreen}{ + 0.7 } ) 	 & 100.0 ( \textcolor{ForestGreen}{ + 0.7 } ) 	 & 99.5 ( \textcolor{ForestGreen}{ + 0.2 } ) 	 & 100.0 ( \textcolor{ForestGreen}{ + 0.7 } ) \\
		60 & staggering                               	 & 99.6 	 & 99.6 ( \textcolor{ForestGreen}{ + 0.0 } ) 	 & 99.6 ( \textcolor{ForestGreen}{ + 0.0 } ) 	 & 96.8 ( \textcolor{red}{  -2.8 } ) 	 & 100.0 ( \textcolor{ForestGreen}{ + 0.4 } ) \\
		\hline
		& \textbf{ Average                                  } 	 &\textbf{ 89.4} 	 &\textbf{ 92.1 } 	 (\textbf{ \textcolor{ForestGreen}{ + 2.8 }}) 	 &\textbf{ 92.6 } 	 (\textbf{ \textcolor{ForestGreen}{ + 3.2 }}) 	 &\textbf{ 94.3 } 	 (\textbf{ \textcolor{ForestGreen}{ + 4.9 }}) 	 &\textbf{ 94.2 } 	 (\textbf{ \textcolor{ForestGreen}{ + 4.6 }}) \\
		\bottomrule
	\end{tabular}
	\vspace{-10pt}
\end{table*}
Table \ref{table_ntu} shows the class-wise performance comparison between our models using Conv-stream only (Model A), using Graph-stream only (Model B), using Conv+Graph streams (Model C), using Conv+Graph streams with shallow layers (Model D), and the baseline (DGNN model of \cite{shi2019skeleton}) on the NTU60 dataset \cite{shahroudy2016ntu}. The comparison shows that our models consistently improve the performance of most of the action classes compared to the baselines. For instance, the models using only Conv-stream (Model A) and only Graph-stream (Model B) produced improvements of 3.8\% and 3.9\% in the average accuracy, respectively compared to the baselines. The combination of Conv and Graph streams (Model C) produced the best performance with improvements of 5.2\% in the average accuracy compared to the baselines. Table \ref{table_ntu} also shows that the lite version of our DeepActsNet which uses shallow layers (Model D) produced improvements of upto 4\% in the average accuracy costing only 3.5 GLOPs compared to the DGNN model of \cite{shi2019skeleton} which costs upto 126 GFLOPs. These improvements are attributed to our Deep Action Stamps which provide highly effective encoding of actions from videos using spatial and motion information from face, hands, and body. Furthermore, our ECGN classifiers learn convolutional and structural features from Deep Action Stamps and produce highly discriminative feature representations which improve the recognition accuracy of most of the action classes.
\subsubsection{Complexity of DeepActsNet}
Table \ref{table_ablation} shows the complexity of DeepActsNet in terms of number of parameters, number of GFlops, and inference time. The results show that the Conv-stream of DeepActsNet costs only 0.8 GFlops compared to the Graph-stream which costs 32.6 GFlops. On the other hand, the Graph-stream has notably small number of parameters and faster inference speed compared to Conv-stream (see row-F in Table \ref{table_ablation} and row-L in Table \ref{table_ablation}).
We also constructed two light versions of our model: DeepActsNet-lite and DeepActsNet-tiny. DeepActsNet-lite uses separate modality-specific branches in the Conv stream and only one branch in the Graph stream. This results in considerable reduction of GLOPS from 33.4 to 3.5 at the cost of small drop in the recognition accuracy as shown in Table \ref{table_ablation} (row-R). DeepActsNet-tiny reduces the number of output channels of all the MBConv blocks in the Conv-stream by a factor of two. This results in a network with only 19.4M parameters. It costs 3.3 GFlops and 62 ms in the forward pass as shown in Table \ref{table_ablation} (row-S). See supplementary material for more details about the model architecture.
\subsubsection{Details of Model Components}
Table \ref{table_network} shows a comparison of the architectures of different variants of the proposed Enhanced Convolutional Graph Network (ECGN). In this work, we developed three variants: 1) A deep version (ECGN - Deep), which consists of a Conv-branch with a configuration (4, 7, 7, 10, 10, 13, 4), representing the number of MBConv layers, and a Graph-branch with a configuration (64, 128, 256), representing the number of output feature channels of the GCN blocks. ECGN-Deep contains 67 M parameters and costs 3.3 GFLOPs.
2) A lite version (ECGN - Lite), which reduces the output feature channels of the GCN blocks of ECGN-Deep by half. This considerably reduces the number of GFLOPs to 0.7 with 64.7 M parameters.
3) A tiny version (ECGN - Tiny), which uses a Conv-branch with a shallow configuration (2, 2, 2, 3, 3, 1, 1) representing the number of MBConv layers, and a Graph-branch with half the number of output feature channels of the GCN blocks. By using shallow configurations in both the Conv and Graph branches, the total parameters of the ECGN-Tiny reduces to only 4.9 M with 0.6 GFLOPs, making it suitable for low-memory applications.
\subsubsection{Comparisons with state-of-the-art Methods}
\begin{table}[htbp]
	\caption{Comparisons of the Cross-Subject (CS) and Cross-View (CV) recognition accuracy with the state-of-the-art methods on the NTU60 dataset \cite{shahroudy2016ntu}.}
	\vspace{-10pt}
	\label{table_sota_ntu60}
	\centering
	\setlength\tabcolsep{3.0pt}
	\begin{tabular}{@{}lccccccccc@{}}
		\toprule
		{Method} &CS(\%)&CV(\%)&Flops \\					
		\midrule
		\rowcolor{grey1}
		\multicolumn{4}{c}{LSTM-based methods}\\
		Deep LSTM \cite{shahroudy2016ntu} &60.7&67.3&-\\
		ST-LSTM \cite{liu2016spatio} &69.2&77.7&-\\
		STA-LSTM \cite{song2016end} &73.4&81.2&-\\
		VA-LSTM \cite{zhang2017view} &79.2&87.7&-\\
		ARRN-LSTM \cite{zheng2018skeleton} &80.7&88.8&-\\
		\hline
		\rowcolor{grey1}
		\multicolumn{4}{c}{CNN-based methods}\\
		HCN \cite{li2018co}  &86.5&91.1&-\\
		TCN \cite{kim2017interpretable} &74.3&83.1&-\\
		Clips+CNN+MTLN \cite{ke2017new} &79.6&84.8&-\\
		CNN \cite{liu2017enhanced}&80.0&87.2&-\\
		ResNet152 \cite{li2017skeleton} &85.0&92.3&-\\
		\hline
		\rowcolor{grey1}
		\multicolumn{4}{c}{Graph-based methods}\\
		ST-GCN \cite{yan2018spatial}  &81.5&88.3&-\\
		GR-GCN \cite{gao2019optimized} &87.5&94.3&-\\
		AS-GCN \cite{li2019actional} &86.8&94.2&27.0\\
		2s-AGCN \cite{shi2019two} &88.5&95.1&35.8\\
		AGC-LSTM \cite{si2019attention} &89.2&95.0&54.4\\
		View-adaptive \cite{zhang2019view} &89.4&95.0&5.4\\
		DGNN \cite{shi2019skeleton}&89.9&96.1&126.8\\
		Shift-GCN \cite{cheng2020skeleton}&90.7&96.5&10.0\\
		MS-G3D Net \cite{liu2020disentangling} &91.5&96.2&19.6\\
		\hline
		DeepActsNet (4s Body Conv+Graph) &\textbf{91.8}&{95.7}&26.8\\		
		DeepActsNet (5s Multi-modal Conv-only) &\textbf{92.1}&{96.4}&0.8\\
	    DeepActsNet (5s Multi-modal Graph-only) &\textbf{92.6}&{95.6}&32.6\\
		DeepActsNet (5s Multi-modal Conv+Graph)&\textbf{94.3}&\textbf{97.5}&33.4\\
		DeepActsNet-lite (5s Multi-modal Conv+Graph) & \textbf{94.2}&\textbf{97.3}&7.1\\
		\bottomrule
	\end{tabular}
\end{table}
\begin{table}[htbp]
	\caption{Comparisons of the Cross-Subject (CS) and Cross-Setup (CSet) recognition accuracy with the state-of-the-art methods on the NTU120 dataset \cite{liu2019ntu}.}
	\vspace{-10pt}
	\label{table_sota_ntu120}
	\centering
	\setlength\tabcolsep{1.5pt}
	\begin{tabular}{@{}lcccccccc@{}}
		\toprule
		{Method} &CS(\%)&CSet(\%)&Flops \\					
		\midrule
		HCN \cite{li2018co}  &78.0&79.8&-\\
		ST-GCN \cite{yan2018spatial}  &73.9&75.9&-\\
		ST-LSTM \cite{liu2017skeleton} &55.7&57.9&-\\
		GCA-LSTM \cite{liu2017global} &61.2&63.3&-\\
		RotClips+MTCNN \cite{ke2018learning} &62.2&61.8&-\\
		Pose Evolution Map \cite{liu2018recognizing} &64.6&66.9&-\\
		2s-AGCN \cite{shi2019two} &82.9&84.9&35.8\\
		Shift-GCN \cite{cheng2020skeleton}&85.9&87.6&10.0\\
		MS-G3D Net \cite{liu2020disentangling} &86.9&88.4&19.6\\
		\hline
		DeepActsNet (4s Body Conv+Graph) &\textbf{87.1}&\textbf{88.8}&26.8\\
		DeepActsNet (5s Multi-modal Conv-only) &{86.9}&\textbf{88.8}&0.8\\
		DeepActsNet (5s Multi-modal Graph-only) &\textbf{87.7}&\textbf{89.4}&32.6\\
		DeepActsNet (5s Multi-modal Conv+Graph)&\textbf{90.2}&\textbf{91.8}&33.4\\
		DeepActsNet-lite (5s Multi-modal Conv+Graph)& \textbf{89.7}&\textbf{91.6}&7.1\\
		\bottomrule
	\end{tabular}
\end{table}
\begin{table}[htbp]
	\caption{Comparisons of the Cross-Subject (CS) and Same-Subject (SS) recognition accuracy with the state-of-the-art methods on the SYSU dataset \cite{hu2015jointly}.}
	\vspace{-10pt}
	\label{table_sota_sysu}
	\centering
	\setlength\tabcolsep{3.0pt}
	\begin{tabular}{@{}lcccccccc@{}}
		\toprule
		{Method} &CS (\%)&SS (\%) \\					
		\midrule
		HCN \cite{li2018co}  &78.4&77.9\\
		ST-GCN \cite{yan2018spatial}  &68.3&66.4\\
		VA-LSTM \cite{zhang2017view} &77.5&76.9\\
		ST-LSTM \cite{liu2017skeleton} &76.5&-\\
		GR-GCN \cite{gao2019optimized} &77.9&-\\
		GCA-LSTM \cite{liu2017skeleton} &78.6&-\\
		SR-TSL \cite{si2018skeleton} &81.9&80.7\\
		GRU \cite{zhang2018adding} &85.7&85.7\\
		View-adaptive \cite{zhang2019view} &86.7&86.2\\
		SGN \cite{zhang2020semantics} &90.6&89.3\\
		\hline
		DeepActsNet (5s Multi-modal Graph-only) &\textbf{91.4}&{84.9}\\
		DeepActsNet (5s Multi-modal Conv-only) &\textbf{92.8}&{88.5}\\
		DeepActsNet (5s Multi-modal Conv+Graph)&\textbf{94.2}&\textbf{90.5}\\
		DeepActsNet-lite (5s Multi-modal Conv+Graph)& \textbf{93.4}&88.9\\
		\bottomrule
	\end{tabular}
\end{table}
Table \ref{table_sota_ntu60}, Table \ref{table_sota_ntu120}, and Table \ref{table_sota_sysu} show the recognition accuracy of our models and the results of state-of-the-art methods on NTU60, NTU120, and SYSU datasets respectively. 
First, we compare our DeepActsNet using only body joints information with the state-of-the-art methods on the NTU60 and NTU120 datasets. Existing methods like MS-G3D \cite{liu2020disentangling} and \cite{cheng2020skeleton} report their best results using an ensemble of 4 data streams (joints, bones, difference of joints between adjacent frames, and difference of bones between adjacent frames). For a fair comparison, we evaluate a variant of our model using the same 4 data streams and the results are represented by (4s Body Conv+Graph) in Table \ref{table_sota_ntu60} and Table \ref{table_sota_ntu120}. The results show that our model produces improvements of 0.3\% and 1.0\% in the CS setting on the NTU60 dataset compared to the methods of MSG-3D and Shift-GCN, respectively. On the NTU120 dataset, our Body-only model produces improvements of 0.2\% and 0.4\% in CS and CSet settings compared to the MS-G3D model, respectively and improvements of 1.2\% compared to the Shift-GCN model, respectively (as shown in Table \ref{table_sota_ntu120}). 
It is also to be noted that our model uses basic graph convolutions (inspired from older ST-GCN \cite{yan2018spatial}) compared to the MS-G3D or DGNN methods that have higher capabilities to model spatial and temporal dependencies between sequential data. Since, the main focus of our paper is to showcase the benefits of multi-modal feature learning for action recognition, we also compared the results of our multi-modal models in Table \ref{table_sota_ntu60}, Table \ref{table_sota_ntu120}, and Table \ref{table_sota_sysu} which beat all the existing methods on the benchmarked datasets with considerable margins.
For instance, DeepActsNet with only Conv-stream produced improvements of around 1.6\% and 2.4\% in the cross-subject and cross-view accuracy on NTU60 dataset compared to the best results of \cite{liu2020disentangling}. 
Furthermore, DeepActsNet with only Graph-stream produced improvements of around 1.8\% and 1.0\% in the cross-subject and cross-view accuracy on the NTU60 dataset compared to the best results of \cite{liu2020disentangling}. 
Table \ref{table_sota_ntu60} also shows that DeepActsNet with Conv+Graph streams yielded improvements of around 3.2\% and 2.8\% compared to the state-of-the-art on the NTU60 dataset. On the SYSU dataset, DeepActsNet produced improvements of around 4.8\% and 5.9\% in the cross-subject and same-subject recognition accuracy compared to the best results of \cite{zhang2020semantics}.
We attribute these improvements to our two innovations: 1) The encoding of spatial and motion information from face, hands, and body in DeepActs capture rich and discriminative cues from videos, resulting in more accurate recognition of action classes involving hand movements and facial deformations compared to the body joints information alone which contain limited information for the face and hands. 2) The modality-specific convolutional and graph layers of our DeepActsNet promote diversity in feature learning and the ensembling reduces variance in the final predictions and improves the action recognition performance.
Table \ref{table_sota_ntu60} and Table \ref{table_sota_ntu120} also show the computational complexity of our models compared to other methods. The results show that DeepActsNet with only Conv-stream achieves higher accuracy than the 4-stream DGNN \cite{shi2019skeleton} and Shift-GCN \cite{cheng2020skeleton} methods with $158\times$ and $12\times$ less GFlops, respectively. DeepActsNet with Conv+Graph streams clearly exceeds all previoulsy reported methods in terms of recognition accuracy with $3.7\times$ less computational cost compared to the 4-stream DGNN \cite{shi2019skeleton}. Compared to Shift-GCN \cite{cheng2020skeleton} and MS-G3D Net \cite{liu2020disentangling}, our DeepActsNet costs 23.4 and 13.8 more GFlops, respectively but produces higher accuracy on NTU60 and NTU120 datasets. Finally, our DeepActsNet-lite presents the best trade-off between computational complexity and recognition accuracy. It clearly wins over the stat-of-the-art for all datasets in terms of recognition accuracy and computational complexity.
%
%
The proposed method is developed for general action recognition from videos and not confined to only skeleton-based action recognition. Therefore, we also compared our method with video-based methods on the NTU and SYSU datasets. For instance, the works of Clips+CNN+MTLN \cite{ke2017new} shown in Table \ref{table_sota_ntu60} and Pose Evolution Map \cite{liu2018recognizing} shown in Table \ref{table_sota_ntu120} used visual cues from RGB image frames in addition to skeleton data. Our method beats these methods with considerable margins.
\section{Conclusion}
In this work, we present two methods for improving action recognition: 1) ``Deep Action Stamps", a multi-modal data representation which encode actions from videos in terms of spatial and motion information from body, face, and hands. 2) ``DeepActsNet", a deep learning based model which learns convolutional and structural features using visual and graph-based representations of Deep Action Stamps for action classification. By coupling these methods, we develop a powerful framework that captures highly informative action cues from videos and learns highly discriminative features for action recognition. Experiments on three public datasets show that our models outperform state-of-the-art methods on all datasets by considerable margins. 
Notably, our work is the first to model spatial and temporal dependencies for body, face, and hands, for action recognition, and our extensive experimental evaluations signify the effectiveness of our approach. 
In future, we plan to enhance the capacity of our model through the integration of attention mechanisms and graph isomorphism layers that have higher expressive capabilities to model spatial and temporal dependecies between sequential joints data. 
We also plan to open source the DeepActsNet model and the Deep Action Stamps data for public research usage. 
\bibliographystyle{IEEEtran}
\bibliography{egbib}

\begin{thebibliography}{10}
\providecommand{\url}[1]{#1}
\csname url@samestyle\endcsname
\providecommand{\newblock}{\relax}
\providecommand{\bibinfo}[2]{#2}
\providecommand{\BIBentrySTDinterwordspacing}{\spaceskip=0pt\relax}
\providecommand{\BIBentryALTinterwordstretchfactor}{4}
\providecommand{\BIBentryALTinterwordspacing}{\spaceskip=\fontdimen2\font plus
\BIBentryALTinterwordstretchfactor\fontdimen3\font minus
  \fontdimen4\font\relax}
\providecommand{\BIBforeignlanguage}[2]{{%
\expandafter\ifx\csname l@#1\endcsname\relax
\typeout{** WARNING: IEEEtran.bst: No hyphenation pattern has been}%
\typeout{** loaded for the language `#1'. Using the pattern for}%
\typeout{** the default language instead.}%
\else
\language=\csname l@#1\endcsname
\fi
#2}}
\providecommand{\BIBdecl}{\relax}
\BIBdecl

\bibitem{fernando2015modeling}
B.~Fernando, E.~Gavves, J.~M. Oramas, A.~Ghodrati, and T.~Tuytelaars,
  ``Modeling video evolution for action recognition,'' in \emph{CVPR}, 2015,
  pp. 5378--5387.

\bibitem{liu2020disentangling}
Z.~Liu, H.~Zhang, Z.~Chen, Z.~Wang, and W.~Ouyang, ``Disentangling and unifying
  graph convolutions for skeleton-based action recognition,'' in
  \emph{Proceedings of the IEEE/CVF Conference on Computer Vision and Pattern
  Recognition}, 2020, pp. 143--152.

\bibitem{shi2019skeleton}
L.~Shi, Y.~Zhang, J.~Cheng, and H.~Lu, ``Skeleton-based action recognition with
  directed graph neural networks,'' in \emph{Proceedings of the IEEE Conference
  on Computer Vision and Pattern Recognition}, 2019, pp. 7912--7921.

\bibitem{si2019attention}
C.~Si, W.~Chen, W.~Wang, L.~Wang, and T.~Tan, ``An attention enhanced graph
  convolutional lstm network for skeleton-based action recognition,'' in
  \emph{Proceedings of the IEEE conference on computer vision and pattern
  recognition}, 2019, pp. 1227--1236.

\bibitem{liu2017skeleton}
J.~Liu, G.~Wang, L.-Y. Duan, K.~Abdiyeva, and A.~C. Kot, ``Skeleton-based human
  action recognition with global context-aware attention lstm networks,''
  \emph{IEEE Transactions on Image Processing}, vol.~27, no.~4, pp. 1586--1599,
  2017.

\bibitem{feichtenhofer2019slowfast}
C.~Feichtenhofer, H.~Fan, J.~Malik, and K.~He, ``Slowfast networks for video
  recognition,'' in \emph{Proceedings of the IEEE international conference on
  computer vision}, 2019, pp. 6202--6211.

\bibitem{caetano2019skelemotion}
C.~Caetano, J.~Sena, F.~Br{\'e}mond, J.~A. Dos~Santos, and W.~R. Schwartz,
  ``Skelemotion: A new representation of skeleton joint sequences based on
  motion information for 3d action recognition,'' in \emph{2019 16th IEEE
  International Conference on Advanced Video and Signal Based Surveillance
  (AVSS)}.\hskip 1em plus 0.5em minus 0.4em\relax IEEE, 2019, pp. 1--8.

\bibitem{shahroudy2016ntu}
A.~Shahroudy, J.~Liu, T.-T. Ng, and G.~Wang, ``Ntu rgb+ d: A large scale
  dataset for 3d human activity analysis,'' in \emph{CVPR}, 2016, pp.
  1010--1019.

\bibitem{liu2019ntu}
J.~Liu, A.~Shahroudy, M.~L. Perez, G.~Wang, L.-Y. Duan, and A.~K. Chichung,
  ``Ntu rgb+ d 120: A large-scale benchmark for 3d human activity
  understanding,'' \emph{PAMI}, 2019.

\bibitem{wang2013learning}
J.~Wang, Z.~Liu, Y.~Wu, and J.~Yuan, ``Learning actionlet ensemble for 3d human
  action recognition,'' \emph{IEEE transactions on pattern analysis and machine
  intelligence}, vol.~36, no.~5, pp. 914--927, 2013.

\bibitem{jin2012essential}
S.-Y. Jin and H.-J. Choi, ``Essential body-joint and atomic action detection
  for human activity recognition using longest common subsequence algorithm,''
  in \emph{Asian Conference on Computer Vision}.\hskip 1em plus 0.5em minus
  0.4em\relax Springer, 2012, pp. 148--159.

\bibitem{hussein2013human}
M.~E. Hussein, M.~Torki, M.~A. Gowayyed, and M.~El-Saban, ``Human action
  recognition using a temporal hierarchy of covariance descriptors on 3d joint
  locations,'' in \emph{Twenty-third international joint conference on
  artificial intelligence}, 2013.

\bibitem{du2015hierarchical}
Y.~Du, W.~Wang, and L.~Wang, ``Hierarchical recurrent neural network for
  skeleton based action recognition,'' in \emph{Proceedings of the IEEE
  conference on computer vision and pattern recognition}, 2015, pp. 1110--1118.

\bibitem{zhu2016co}
W.~Zhu, C.~Lan, J.~Xing, W.~Zeng, Y.~Li, L.~Shen, and X.~Xie, ``Co-occurrence
  feature learning for skeleton based action recognition using regularized deep
  lstm networks,'' in \emph{Thirtieth AAAI Conference on Artificial
  Intelligence}, 2016.

\bibitem{zhang2019view}
P.~Zhang, C.~Lan, J.~Xing, W.~Zeng, J.~Xue, and N.~Zheng, ``View adaptive
  neural networks for high performance skeleton-based human action
  recognition,'' \emph{IEEE transactions on pattern analysis and machine
  intelligence}, vol.~41, no.~8, pp. 1963--1978, 2019.

\bibitem{li2018co}
C.~Li, Q.~Zhong, D.~Xie, and S.~Pu, ``Co-occurrence feature learning from
  skeleton data for action recognition and detection with hierarchical
  aggregation,'' \emph{arXiv preprint arXiv:1804.06055}, 2018.

\bibitem{li2019learning}
Y.~Li, R.~Xia, X.~Liu, and Q.~Huang, ``Learning shape-motion representations
  from geometric algebra spatio-temporal model for skeleton-based action
  recognition,'' in \emph{2019 IEEE International Conference on Multimedia and
  Expo (ICME)}.\hskip 1em plus 0.5em minus 0.4em\relax IEEE, 2019, pp.
  1066--1071.

\bibitem{liu2017enhanced}
M.~Liu, H.~Liu, and C.~Chen, ``Enhanced skeleton visualization for view
  invariant human action recognition,'' \emph{Pattern Recognition}, vol.~68,
  pp. 346--362, 2017.

\bibitem{feichtenhofer2016convolutional}
C.~Feichtenhofer, A.~Pinz, and A.~Zisserman, ``Convolutional two-stream network
  fusion for video action recognition,'' in \emph{Proceedings of the IEEE
  conference on computer vision and pattern recognition}, 2016, pp. 1933--1941.

\bibitem{simonyan2014two}
K.~Simonyan and A.~Zisserman, ``Two-stream convolutional networks for action
  recognition in videos,'' in \emph{Advances in neural information processing
  systems}, 2014, pp. 568--576.

\bibitem{li2019temporal}
Y.~Li, S.~Song, Y.~Li, and J.~Liu, ``Temporal bilinear networks for video
  action recognition,'' in \emph{Proceedings of the AAAI Conference on
  Artificial Intelligence}, vol.~33, 2019, pp. 8674--8681.

\bibitem{jiang2019stm}
B.~Jiang, M.~Wang, W.~Gan, W.~Wu, and J.~Yan, ``Stm: Spatiotemporal and motion
  encoding for action recognition,'' in \emph{Proceedings of the IEEE
  International Conference on Computer Vision}, 2019, pp. 2000--2009.

\bibitem{xie2017aggregated}
S.~Xie, R.~Girshick, P.~Doll{\'a}r, Z.~Tu, and K.~He, ``Aggregated residual
  transformations for deep neural networks,'' in \emph{Proceedings of the IEEE
  conference on computer vision and pattern recognition}, 2017, pp. 1492--1500.

\bibitem{qiu2017learning}
Z.~Qiu, T.~Yao, and T.~Mei, ``Learning spatio-temporal representation with
  pseudo-3d residual networks,'' in \emph{ICCV}, 2017, pp. 5533--5541.

\bibitem{li2019spatio}
B.~Li, X.~Li, Z.~Zhang, and F.~Wu, ``Spatio-temporal graph convolution for
  skeleton-based action recognition,'' in \emph{Proceedings of the AAAI
  Conference on Artificial Intelligence}, vol.~33, 2019, pp. 8561--8568.

\bibitem{li2019actional}
M.~Li, S.~Chen, X.~Chen, Y.~Zhang, Y.~Wang, and Q.~Tian, ``Actional-structural
  graph convolutional networks for skeleton-based action recognition,'' in
  \emph{Proceedings of the IEEE Conference on Computer Vision and Pattern
  Recognition}, 2019, pp. 3595--3603.

\bibitem{shi2019two}
L.~Shi, Y.~Zhang, J.~Cheng, and H.~Lu, ``Two-stream adaptive graph
  convolutional networks for skeleton-based action recognition,'' in
  \emph{Proceedings of the IEEE Conference on Computer Vision and Pattern
  Recognition}, 2019, pp. 12\,026--12\,035.

\bibitem{gao2019optimized}
X.~Gao, W.~Hu, J.~Tang, J.~Liu, and Z.~Guo, ``Optimized skeleton-based action
  recognition via sparsified graph regression,'' in \emph{Proceedings of the
  27th ACM International Conference on Multimedia}, 2019, pp. 601--610.

\bibitem{liu2019skepxels}
J.~Liu, N.~Akhtar, and A.~Mian, ``Skepxels: Spatio-temporal image
  representation of human skeleton joints for action recognition.'' in
  \emph{CVPR Workshops}, 2019.

\bibitem{hu2015jointly}
J.-F. Hu, W.-S. Zheng, J.~Lai, and J.~Zhang, ``Jointly learning heterogeneous
  features for rgb-d activity recognition,'' in \emph{Proceedings of the IEEE
  conference on computer vision and pattern recognition}, 2015, pp. 5344--5352.

\bibitem{wu2019detectron2}
Y.~Wu, A.~Kirillov, F.~Massa, W.-Y. Lo, and R.~Girshick, ``Detectron2,''
  \url{https://github.com/facebookresearch/detectron2}, 2019.

\bibitem{simon2017hand}
T.~Simon, H.~Joo, I.~Matthews, and Y.~Sheikh, ``Hand keypoint detection in
  single images using multiview bootstrapping,'' in \emph{CVPR}, 2017, pp.
  1145--1153.

\bibitem{sagonas2016300}
C.~Sagonas, E.~Antonakos, G.~Tzimiropoulos, S.~Zafeiriou, and M.~Pantic, ``300
  faces in-the-wild challenge: Database and results,'' \emph{Image and vision
  computing}, vol.~47, pp. 3--18, 2016.

\bibitem{deng2019retinaface}
J.~Deng, J.~Guo, Y.~Zhou, J.~Yu, I.~Kotsia, and S.~Zafeiriou, ``Retinaface:
  Single-stage dense face localisation in the wild,'' \emph{arXiv preprint
  arXiv:1905.00641}, 2019.

\bibitem{tan2019efficientnet}
M.~Tan and Q.~V. Le, ``Efficientnet: Rethinking model scaling for convolutional
  neural networks,'' \emph{arXiv preprint arXiv:1905.11946}, 2019.

\bibitem{sandler2018mobilenetv2}
M.~Sandler, A.~Howard, M.~Zhu, A.~Zhmoginov, and L.-C. Chen, ``Mobilenetv2:
  Inverted residuals and linear bottlenecks,'' in \emph{CVPR}, 2018, pp.
  4510--4520.

\bibitem{hu2018squeeze}
J.~Hu, L.~Shen, and G.~Sun, ``Squeeze-and-excitation networks,'' in
  \emph{CVPR}, 2018, pp. 7132--7141.

\bibitem{yan2018spatial}
S.~Yan, Y.~Xiong, and D.~Lin, ``Spatial temporal graph convolutional networks
  for skeleton-based action recognition,'' in \emph{AAAI}, 2018.

\bibitem{liu2016spatio}
J.~Liu, A.~Shahroudy, D.~Xu, and G.~Wang, ``Spatio-temporal lstm with trust
  gates for 3d human action recognition,'' in \emph{ECCV}.\hskip 1em plus 0.5em
  minus 0.4em\relax Springer, 2016, pp. 816--833.

\bibitem{song2016end}
S.~Song, C.~Lan, J.~Xing, W.~Zeng, and J.~Liu, ``An end-to-end spatio-temporal
  attention model for human action recognition from skeleton data,''
  \emph{arXiv preprint arXiv:1611.06067}, 2016.

\bibitem{zhang2017view}
P.~Zhang, C.~Lan, J.~Xing, W.~Zeng, J.~Xue, and N.~Zheng, ``View adaptive
  recurrent neural networks for high performance human action recognition from
  skeleton data,'' in \emph{Proceedings of the IEEE International Conference on
  Computer Vision}, 2017, pp. 2117--2126.

\bibitem{zheng2018skeleton}
W.~Zheng, L.~Li, Z.~Zhang, Y.~Huang, and L.~Wang, ``Skeleton-based relational
  modeling for action recognition,'' \emph{arXiv preprint arXiv:1805.02556},
  2018.

\bibitem{kim2017interpretable}
T.~S. Kim and A.~Reiter, ``Interpretable 3d human action analysis with temporal
  convolutional networks,'' in \emph{2017 IEEE conference on computer vision
  and pattern recognition workshops (CVPRW)}.\hskip 1em plus 0.5em minus
  0.4em\relax IEEE, 2017, pp. 1623--1631.

\bibitem{ke2017new}
Q.~Ke, M.~Bennamoun, S.~An, F.~Sohel, and F.~Boussaid, ``A new representation
  of skeleton sequences for 3d action recognition,'' in \emph{Proceedings of
  the IEEE conference on computer vision and pattern recognition}, 2017, pp.
  3288--3297.

\bibitem{li2017skeleton}
B.~Li, Y.~Dai, X.~Cheng, H.~Chen, Y.~Lin, and M.~He, ``Skeleton based action
  recognition using translation-scale invariant image mapping and multi-scale
  deep cnn,'' in \emph{2017 IEEE International Conference on Multimedia \& Expo
  Workshops (ICMEW)}.\hskip 1em plus 0.5em minus 0.4em\relax IEEE, 2017, pp.
  601--604.

\bibitem{cheng2020skeleton}
K.~Cheng, Y.~Zhang, X.~He, W.~Chen, J.~Cheng, and H.~Lu, ``Skeleton-based
  action recognition with shift graph convolutional network,'' in \emph{CVPR},
  2020, pp. 183--192.

\bibitem{liu2017global}
J.~Liu, G.~Wang, P.~Hu, L.-Y. Duan, and A.~C. Kot, ``Global context-aware
  attention lstm networks for 3d action recognition,'' in \emph{Proceedings of
  the IEEE Conference on Computer Vision and Pattern Recognition}, 2017, pp.
  1647--1656.

\bibitem{ke2018learning}
Q.~Ke, M.~Bennamoun, S.~An, F.~Sohel, and F.~Boussaid, ``Learning clip
  representations for skeleton-based 3d action recognition,'' \emph{IEEE
  Transactions on Image Processing}, vol.~27, no.~6, pp. 2842--2855, 2018.

\bibitem{liu2018recognizing}
M.~Liu and J.~Yuan, ``Recognizing human actions as the evolution of pose
  estimation maps,'' in \emph{CVPR}, 2018, pp. 1159--1168.

\bibitem{si2018skeleton}
C.~Si, Y.~Jing, W.~Wang, L.~Wang, and T.~Tan, ``Skeleton-based action
  recognition with spatial reasoning and temporal stack learning,'' in
  \emph{ECCV}, 2018, pp. 103--118.

\bibitem{zhang2018adding}
P.~Zhang, J.~Xue, C.~Lan, W.~Zeng, Z.~Gao, and N.~Zheng, ``Adding attentiveness
  to the neurons in recurrent neural networks,'' in \emph{ECCV}, 2018, pp.
  135--151.

\bibitem{zhang2020semantics}
P.~Zhang, C.~Lan, W.~Zeng, J.~Xing, J.~Xue, and N.~Zheng, ``Semantics-guided
  neural networks for efficient skeleton-based human action recognition,'' in
  \emph{CVPR}, 2020, pp. 1112--1121.

\bibitem{cheron2015p}
G.~Ch{\'e}ron, I.~Laptev, and C.~Schmid, ``P-cnn: Pose-based cnn features for
  action recognition,'' in \emph{Proceedings of the IEEE international
  conference on computer vision}, 2015, pp. 3218--3226.

\bibitem{inproceedings}
S.~Das, A.~Chaudhary, and M.~Thonnat, ``Where to focus on for human action
  recognition?'' 01 2019, pp. 71--80.

\bibitem{baradel2017pose}
F.~Baradel, C.~Wolf, and J.~Mille, ``Pose-conditioned spatio-temporal attention
  for human action recognition,'' \emph{arXiv preprint arXiv:1703.10106}, 2017.

\bibitem{garcia2018first}
G.~Garcia-Hernando, S.~Yuan, S.~Baek, and T.-K. Kim, ``First-person hand action
  benchmark with rgb-d videos and 3d hand pose annotations,'' in
  \emph{Proceedings of the IEEE conference on computer vision and pattern
  recognition}, 2018, pp. 409--419.

\bibitem{lei2012fine}
J.~Lei, X.~Ren, and D.~Fox, ``Fine-grained kitchen activity recognition using
  rgb-d,'' in \emph{Proceedings of the 2012 ACM Conference on Ubiquitous
  Computing}, 2012, pp. 208--211.

\bibitem{li2017action}
W.~Li, F.~Abtahi, and Z.~Zhu, ``Action unit detection with region adaptation,
  multi-labeling learning and optimal temporal fusing,'' in \emph{Proceedings
  of the IEEE Conference on Computer Vision and Pattern Recognition}, 2017, pp.
  1841--1850.

\end{thebibliography}
\end{document}